\documentclass[nohyperref]{article}

\usepackage{microtype}
\usepackage{graphicx}
\usepackage{subfig}
\usepackage{booktabs} % for professional tables
\usepackage{amsmath}
\usepackage{amssymb}
\usepackage{booktabs}
\usepackage{pifont}
\usepackage[skip=1pt]{caption}
\usepackage[table]{xcolor}
\usepackage{array}
\usepackage{multirow}

\definecolor{citecolor}{HTML}{0071BC}
\definecolor{linkcolor}{HTML}{ED1C24}

\usepackage[pagebackref=false, breaklinks=true, letterpaper=true, colorlinks, citecolor=citecolor, linkcolor=linkcolor, bookmarks=false]{hyperref}

\usepackage[accepted]{arxiv}

\usepackage{amsmath}
\usepackage{amssymb}
\usepackage{mathtools}
\usepackage{amsthm}

\usepackage[capitalize,noabbrev]{cleveref}

\crefname{section}{Sec.}{Secs.}
\Crefname{section}{Section}{Sections}
\Crefname{table}{Table}{Tables}
\crefname{table}{Tab.}{Tabs.}

\newlength\savewidth\newcommand\shline{\noalign{\global\savewidth\arrayrulewidth
		\global\arrayrulewidth 1pt}\hline\noalign{\global\arrayrulewidth\savewidth}}
\newcommand{\tablestyle}[2]{\setlength{\tabcolsep}{#1}\renewcommand{\arraystretch}{#2}\centering\footnotesize}
\renewcommand{\paragraph}[1]{\vspace{1.25mm}\noindent\textbf{#1}}

\newcolumntype{x}[1]{>{\centering\arraybackslash}p{#1pt}}
\newcolumntype{y}[1]{>{\raggedright\arraybackslash}p{#1pt}}
\newcolumntype{z}[1]{>{\raggedleft\arraybackslash}p{#1pt}}

\definecolor{baselinecolor}{gray}{.9}
\newcommand{\baseline}[1]{\cellcolor{baselinecolor}{#1}}
\definecolor{deemph}{gray}{0.6}
\newcommand{\gc}[1]{\textcolor{deemph}{#1}}
\newcommand{\ul}[1]{\underline{#1}}

\definecolor{ourscolor}{gray}{0.9}
\newcommand{\ours}[1]{\cellcolor{ourscolor}{#1}}

\definecolor{imgcolor}{HTML}{2684F7}
\definecolor{vidcolor}{HTML}{F03C4E}

\newcommand{\img}[1]{\textcolor{imgcolor}{#1}}
\newcommand{\vid}[1]{\textcolor{vidcolor}{#1}}

\newcommand{\chkmark}{\text{\ding{52}}}%
\newcommand{\xmark}{\text{\ding{55}}}%
\newcommand{\x}{{$\times$}}

\newcommand{\fullname}{Simple Hierarchical Vision Transformer}
\newcommand{\name}{{Hiera}}
\newcommand{\shortname}{{Hiera}}

\theoremstyle{plain}

\theoremstyle{definition}

\theoremstyle{remark}

\newcommand{\boxAP}{AP$^\text{box}$~}
\newcommand{\maskAP}{AP$^\text{mask}$~}
\newcommand{\flops}[3]{#1$\times$#2$\times$#3}

\usepackage[textsize=tiny]{todonotes}

\icmltitlerunning{Hiera: A Hierarchical Vision Transformer without the Bells-and-Whistles}

\begin{document}

\twocolumn[
\icmltitle{\name: A Hierarchical Vision Transformer without the Bells-and-Whistles}



\icmlsetsymbol{equal}{*}

\begin{icmlauthorlist}
\icmlauthor{Chaitanya Ryali}{equal,meta}
\icmlauthor{Yuan-Ting Hu}{equal,meta}
\icmlauthor{Daniel Bolya}{equal,meta,geor}
\icmlauthor{Chen Wei}{meta,hop}
\icmlauthor{Haoqi Fan}{meta}\\
\icmlauthor{Po-Yao Huang}{meta}
\icmlauthor{Vaibhav Aggarwal}{meta}
\icmlauthor{Arkabandhu Chowdhury}{meta}
\icmlauthor{Omid Poursaeed}{meta}\\
\icmlauthor{Judy Hoffman}{geor}
\icmlauthor{Jitendra Malik}{meta}
\icmlauthor{Yanghao Li}{equal,meta}
\icmlauthor{Christoph Feichtenhofer}{equal,meta}
\end{icmlauthorlist}

\icmlaffiliation{meta}{Meta AI, FAIR}
\icmlaffiliation{geor}{Georgia Tech}
\icmlaffiliation{hop}{Johns Hopkins University}

\icmlcorrespondingauthor{Chaitanya Ryali}{chayryali@meta.com}
% \icmlcorrespondingauthor{Christoph Feichtenhofer}{feichtenhofer@meta.com}

\icmlkeywords{Computer Vision, Self-Supervised Learning, Classification, Detection}

\vskip 0.3in
]

\printAffiliationsAndNotice{\icmlEqualContribution} % otherwise use the standard text.

\begin{abstract}
    Modern hierarchical vision transformers have added several vision-specific components in the pursuit of supervised classification performance. While these components lead to effective accuracies and attractive FLOP counts, the added complexity actually makes these transformers \textit{slower} than their vanilla ViT counterparts. In this paper, we argue that this additional bulk is \textit{unnecessary}. By pretraining with a strong visual pretext task (MAE), we can strip out all the bells-and-whistles from a state-of-the-art multi-stage vision transformer \textit{without losing accuracy}. In the process, we create \name{}, an extremely simple hierarchical vision transformer that is \textit{more accurate} than previous models while being \textit{significantly faster} both at inference and during training. We evaluate \name{} on a variety of tasks for image and video recognition. Our code and models are available at \hypersetup{urlcolor=magenta}\href{https://github.com/facebookresearch/hiera}{https://github.com/facebookresearch/hiera}.
\vspace{-5pt}
\end{abstract}

\section{Introduction}

Since their introduction by \citet{vit} a few years ago, Vision Transformers (ViTs) have dominated several tasks in computer vision. While architecturally simple, their accuracy \cite{deit3} and ability to scale \cite{scalingvits} make them still a popular choice today. 
Moreover, their simplicity unlocks the use of powerful pretraining strategies such as MAE \cite{mae}, which make ViTs computationally and data efficient to train.

However, this simplicity comes at a cost: by using the same spatial resolution and number of channels throughout the network, ViTs make inefficient use of their parameters. This is in contrast to prior ``hierarchical'' or ``multi-scale'' models (e.g., \citet{alexnet,resnet}), which use fewer channels but higher spatial resolution in early stages %of the model
with simpler features, and more channels but lower spatial resolution later in the model % at the end of the model
with more complex features.

Several domain specific vision transformers have been introduced that employ this hierarchical design, such as Swin \cite{swin} or MViT \cite{mvitv1}. However, in the pursuit of state-of-the-art results using fully supervised training on ImageNet-1K (an area where ViT has historically struggled), these models have become more and more complicated as they add specialized modules (e.g., cross-shaped windows in CSWin \cite{cswin}, decomposed relative position embeddings in MViTv2 \cite{mvitv2}). While these changes produce effective models with attractive floating point operation (FLOP) counts, under the hood the added complexity makes these models \textit{slower} overall.

%##################################################################################################
\begin{figure}[t!]
    \centering
    \includegraphics[width=\linewidth]{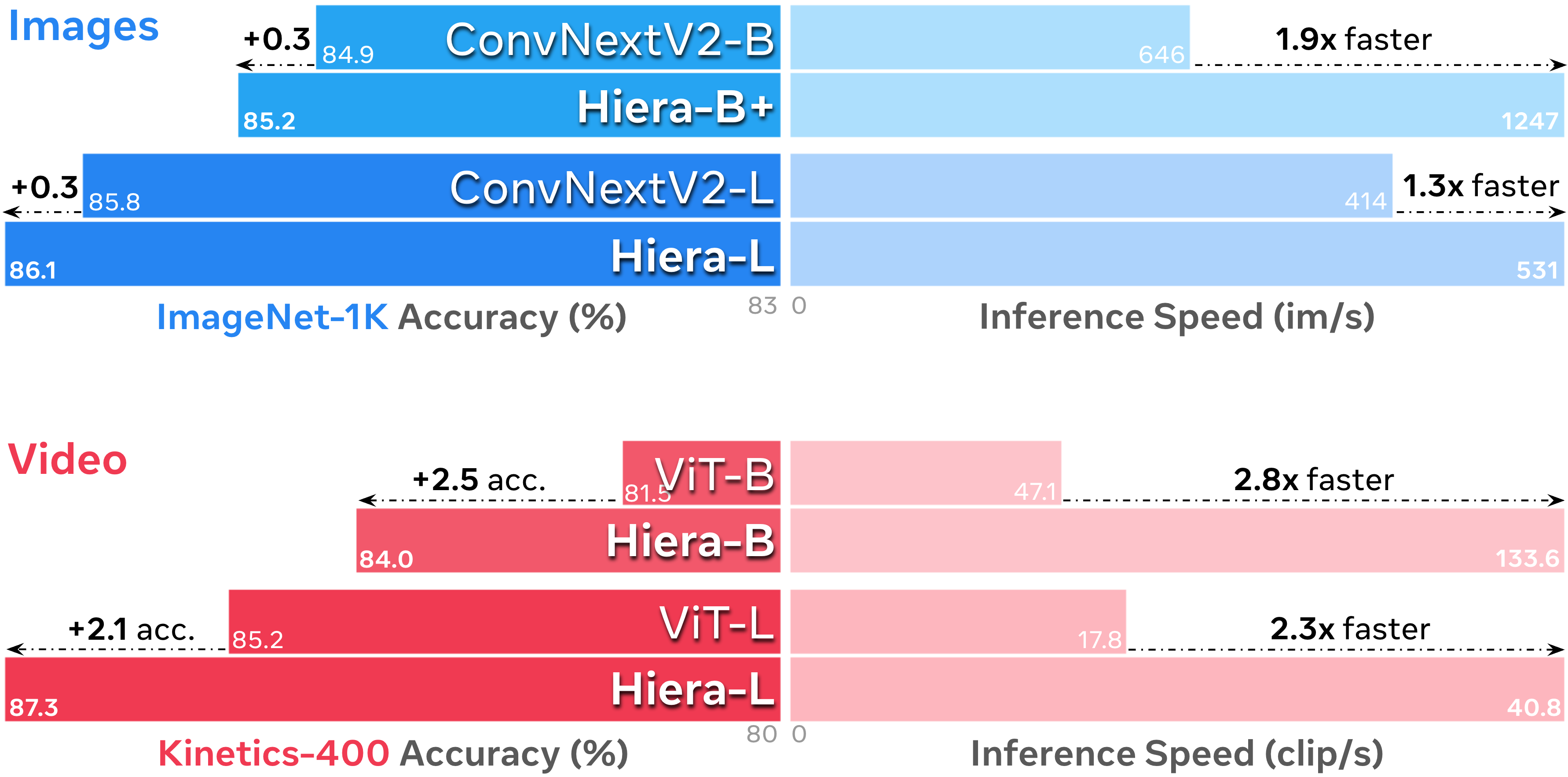}
    \vspace{-15pt}
    \caption{ \textbf{\name{}} cuts out expensive specialized operations (e.g., convs) from hierarchical transformers to create a simple, efficient, and accurate model that is fast across many image and video tasks. Above we compare to recent MAE-based works \cite{woo2023convnextv2,mae-st}. All speeds measured with A100, fp16. % All benchmarks use a single A100 with fp16 unless otherwise noted.
    }
    \label{fig:concept}
    \vspace{-10pt}
\end{figure}
%##################################################################################################
%##################################################################################################
\begin{figure*}[th]
\centering
\includegraphics[width=0.89\linewidth]{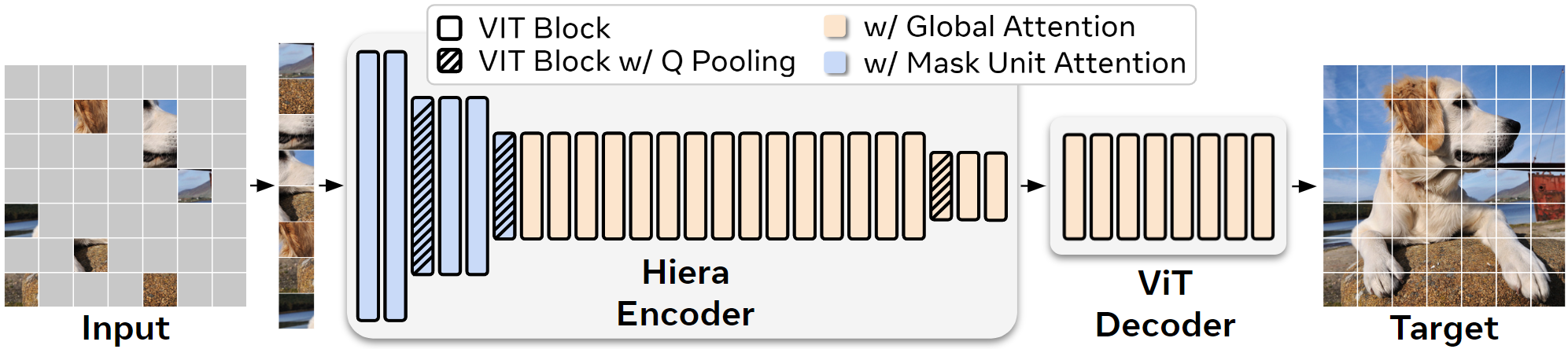}
\vspace{-.1em}
\caption{\textbf{\shortname{} Setup.}
Modern hierarchical transformers like Swin \cite{swin} or MViT \cite{mvitv2} are more parameter efficient than vanilla ViTs \cite{vit}, but end up slower due to overhead from adding spatial bias through vision-specific modules like shifted windows or convs. In contrast, we design \shortname{} to be as simple as possible. To add spatial bias, we opt to \textit{teach} it to the model using a strong pretext task like MAE (pictured here) instead. \shortname{} consists entirely of standard ViT blocks. For efficiency, we use local attention within ``mask units'' (Fig.~\ref{fig:architecture_overlap},~\ref{fig:pool_vs_mu_attn}) for the first two stages and global attention for the rest.
At each stage transition, $Q$ and the skip connection have their features doubled by a linear layer and spatial dimension pooled by a $2\times2$ maxpool. \shortname{-B} is shown here (see Tab.~\ref{tab:hiera_configs} for other configs).
}
\label{fig:arch_and_task}\vspace{.5em}
\end{figure*}
%##################################################################################################

We argue that a lot of this bulk is actually \textit{unnecessary}. 
Because ViTs lack inductive bias after their initial patchify operation, many of the changes proposed by subsequent vision specific transformers serve to manually add spatial biases. But why should we slow down our architecture to add these biases, if we could just train the model to learn them instead? In particular, MAE pretraining has shown to be a very effective tool to teach ViTs spatial reasoning, allowing pure vision transformers to obtain good results on detection \cite{vitdet}, which was a task previously dominated by models like Swin or MViT. Moreover, MAE pretraining is \textit{sparse} and can be $4-10\times$ as fast as normal supervised training, making it an already desirable alternative across many domains for more than just accuracy \cite{mae,mae-st,mae-audio}.

We test this hypothesis with a simple strategy: using some implementation tricks (Fig.~\ref{fig:architecture_overlap}), take an existing hierarchical ViT (e.g., MViTv2) and carefully \textit{remove} non-essential components while training with MAE (Tab.~\ref{tab:sparsifying_mvit}). After tuning the MAE task to this new architecture (Tab.~\ref{tab:ablations}), we find that we can actually simplify or remove \textit{all} of the non-transformer components, while \textit{increasing in accuracy}. The result is an extremely efficient model with no bells-and-whistles: no convolutions, no shifted or cross-shaped windows, no decomposed relative position embeddings. Just a pure, simple hierarchical ViT that is both \textit{faster and more accurate} than prior work across several model sizes, domains, and tasks.

\begin{figure}[t]
    \centering
    \includegraphics[width=1\linewidth]{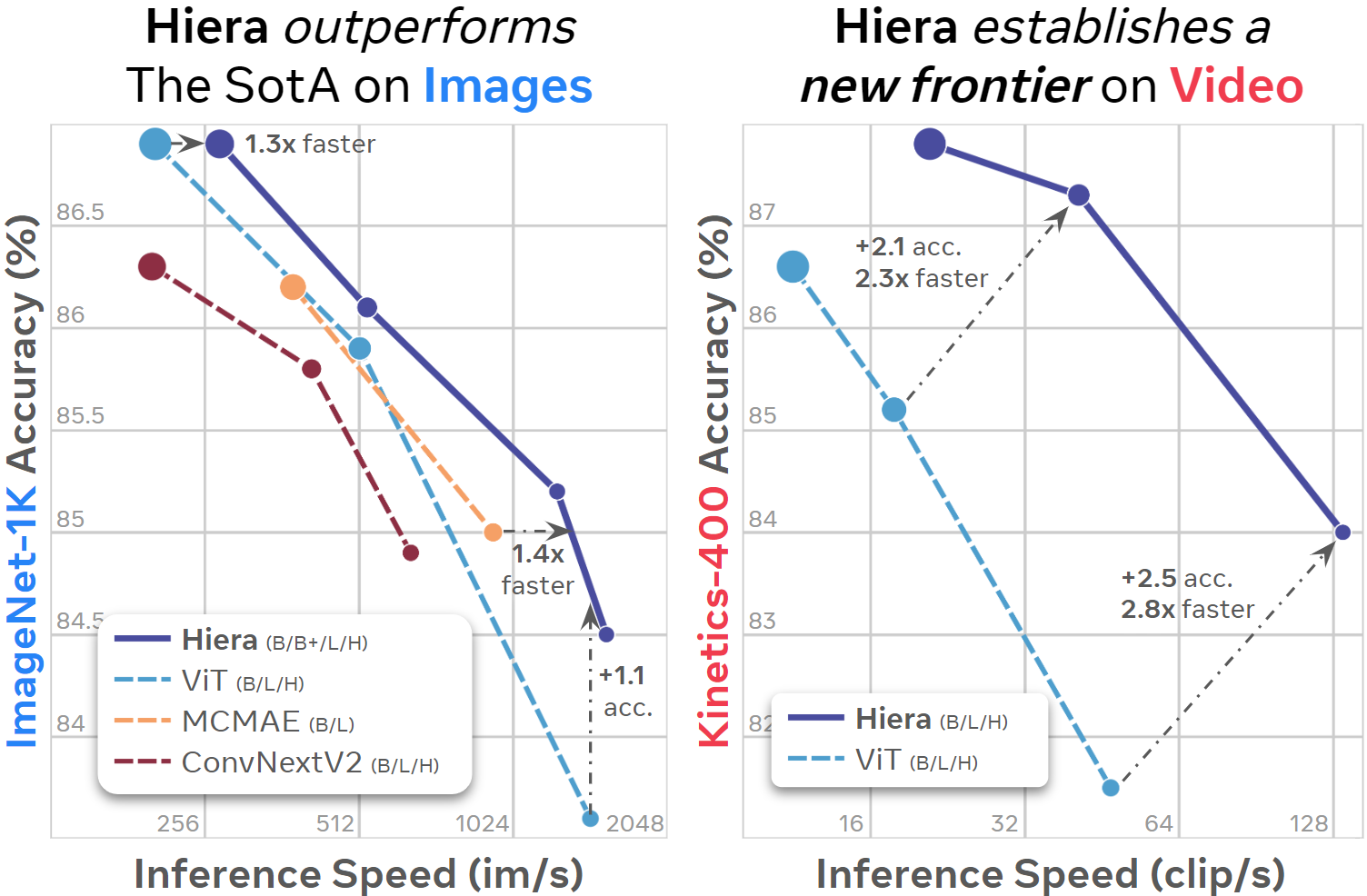}
    \vspace{-2em}
    \caption{\textbf{Performance vs. prior work.} \shortname{} compared to B, L, and H variants of SotA models that use MAE-like pretraining. On \img{images}, \shortname{} is faster and more accurate than even the most recent SotA \cite{mae,gao2022mcmae,woo2023convnextv2}, offering 30-40\% speed-up compared to the best model at every scale. On \vid{video}, \shortname{} represents a new class of performance, significantly improving accuracy, while \textit{being over 2\x~faster} than popular ViT models.
    Marker size is proportional to FLOP count.
    }
    \label{fig:imnet1k_inference_speed}
\end{figure}

Our \fullname{} (\shortname{}) outperforms the SotA on \img{images} and \textit{far exceeds} prior work on \vid{video} while being \textit{much faster} (Fig.~\ref{fig:concept}) at every model scale (Fig.~\ref{fig:imnet1k_inference_speed}) and across extensive datasets and tasks (Sec.~\ref{sec:video_experiments},~\ref{sec:image_experiments}).

\section{Related Work}
\noindent\textbf{Vision transformers (ViTs)} have attracted attention because of their massive success on several vision tasks including image classification~\cite{vit}, video classification~\cite{mvitv1,vivit,bertasius2021space}, semantic segmentation~\cite{ranftl2021vision}, object detection~\cite{carion2020end,vitdet}, video object segmentation~\cite{duke2021sstvos}, 3D object detection~\cite{misra2021end} and 3D reconstruction~\cite{bozic2021transformerfusion}. The key difference between vanilla ViT~\cite{vit} and prior convolutional neural networks (CNNs)~\cite{lecun1998gradient} is that ViT partitions images into, e.g., 16×16 pixel, \emph{non-overlapping} patches and flattens the spatial grid into a 1D sequence, whereas CNNs maintain this grid over multiple stages of the model, reducing the resolution in each stage and introducing inductive biases such as shift equivariance. Recently, the field has shown an increased interest in hybrid methods~\cite{mvitv1,swin,mvitv2,cswin,pvt} that combine transformers with convolution-like operations and the hierarchical stage structure of prior CNNs. This direction has shown success and has achieved state-of-the-art on various vision tasks. However, in practice these models are actually \textit{slower} than their vanilla ViT counterparts and convs are not easily compatible with popular self-supervised tasks such as masked image modeling. We address both of these issues in the creation of \name{}.

%##################################################################################################
%##################################################################################################

\paragraph{Masked pretraining} has emerged as a powerful self-supervised learning pretext task for learning visual representations~\cite{vincent2010stacked,pathak2016context,chen2020generative,mae,bao2021beit,xie2021simmim,milan}. Among previous works, Masked AutoEncoder (MAE, \citet{mae}) takes advantage of vanilla ViTs, which allow any length of input, and thereby derives an efficient training regime using the \emph{sparsity} of masked images. This greatly improves the training efficiency of masked pretraining, but adapting sparse training to hierarchical models is nontrivial, 
because the input is no longer laid out in a rigid 2D grid. 
There have been several attempts to enable hierarchical ViTs to use masked pretraining. MaskFeat~\cite{maskfeat} and SimMIM \cite{xie2021simmim} replace masked patches with \texttt{[mask]} tokens, meaning most computation is wasted on non-visible tokens and training is incredibly slow.
\citet{huang2022green} introduce several techniques to enable sparsity in every component of the network, in the end creating a much more complicated model that doesn't improve much in accuracy.  UM-MAE~\cite{li2022uniform} uses a special masking strategy to allow for sparsity, but this restriction significantly hurts accuracy.
MCMAE~\cite{gao2022mcmae} uses masked convolution in the first couple of stages which obtains high accuracy but significantly reduces the efficiency of the model overall. We bypass all of these complicated techniques and restrictions by designing our architecture specifically for \textit{sparse} MAE pretraining, thereby creating a powerful yet simple model.

\section{Approach} \label{sec:approach}

Our goal is to create a powerful and efficient multiscale vision transformer that is, above all, \textit{simple}. We argue that we do not need any specialized modules like convolution \cite{mvitv1}, shifted windows \cite{swin}, or attention bias \cite{levit,mvitv2} to obtain high accuracy on vision tasks. This may seem difficult, as these techniques add much needed spatial (and temporal) biases that vanilla transformers \cite{vit} lack. However, we employ a different strategy. While prior work adds spatial bias through complicated architectural changes, we opt to keep the model simple and \textit{learn} these biases through a strong pretext task instead. To show the efficacy of this idea, we devise a simple experiment: take an existing hierarchical vision transformer and ablate its bells-and-whistles while training with a strong pretext task.

%##################################################################################################
\begin{figure}[t]
\centering
\includegraphics[width=0.99\linewidth]{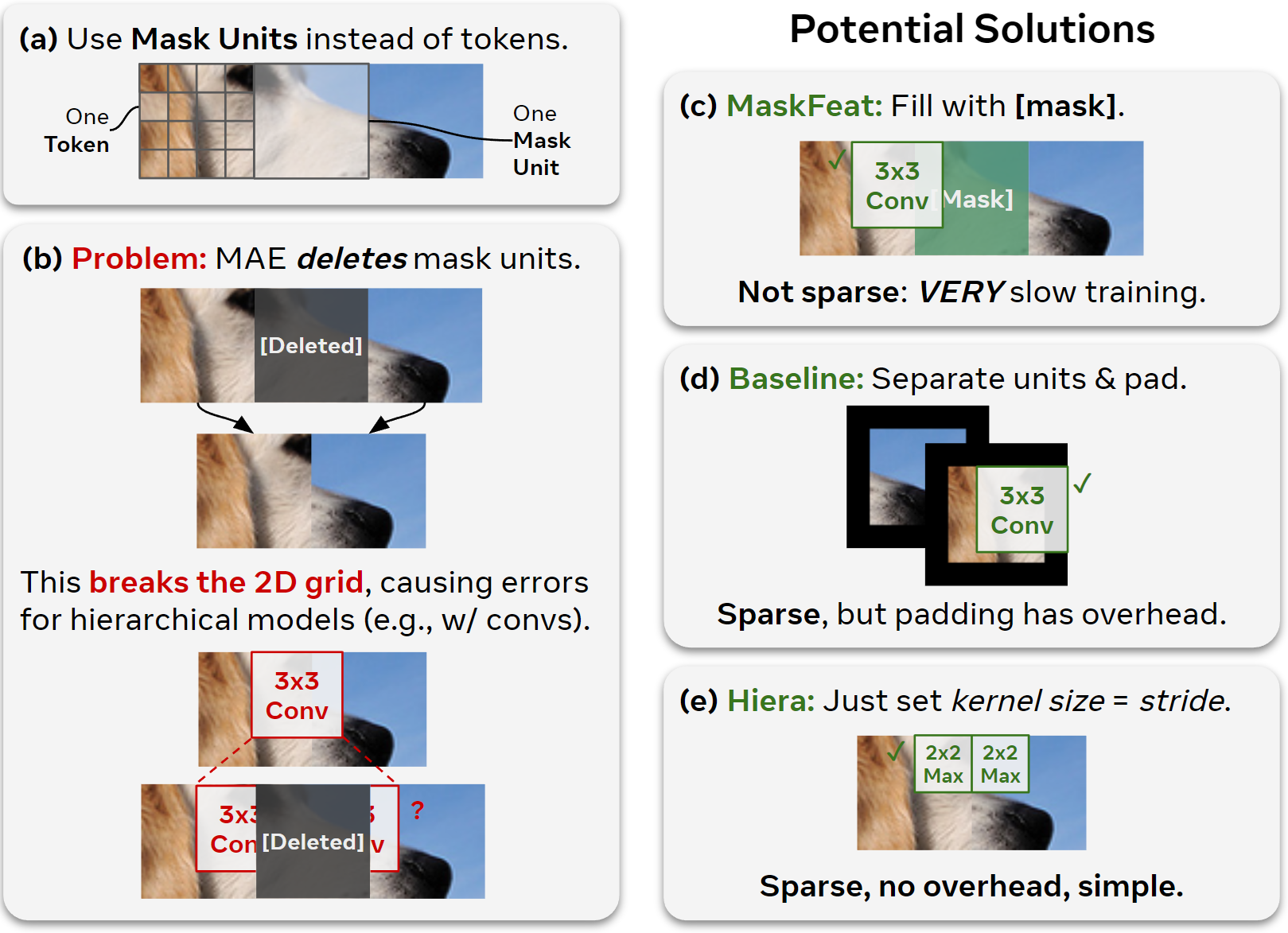}
\vspace{-.3em}

\caption{\textbf{MAE for Hierarchical Models.}
MAE is not compatible with multi-stage models, but we can apply some simple tricks to remedy this.
While MAE masks individual tokens, tokens in multi-stage transformers start very small (e.g., $4\times4$ pixels), doubling size in each stage.
(a) Thus, we mask coarser ``mask units'' ($32\times32$ pixels) instead of tokens directly.
(b) For efficiency, MAE is \textit{sparse}, meaning it \textit{deletes} what it masks (a problem for spatial modules like convs).
(c) Keeping masked tokens fixes this, but gives up the potential $4-10\times$ training speed-up of MAE.
(d) As a baseline, we introduce a trick that treats mask units as a separate entities for convs, solving the issue but requiring undesirable padding.
(e) In \shortname{}, we side-step the problem entirely by changing the architecture so the kernels can't overlap between mask units.
}
\label{fig:architecture_overlap}
\vspace{-1em}
\end{figure}
%##################################################################################################

For the pretext task, we use Masked Autoencoders (MAE, \citet{mae}), which has been shown effective in teaching ViTs localization capabilities for downstream tasks (e.g., detection \cite{vitdet}) by having the network reconstruct masked input patches (Fig.~\ref{fig:arch_and_task}). Note that MAE pretraining is \textit{sparse}---that is, masked tokens are \textit{deleted} instead of being overwritten like in other masked image modeling approaches \cite{maskfeat,xie2021simmim}. This makes pretraining efficient, but poses a problem for existing hierarchical models as it breaks the 2D grid that they rely on (Fig.~\ref{fig:architecture_overlap}\hyperref[fig:architecture_overlap]{b}). Moreover, MAE masks out individual tokens, which are large $16\times16$ patches for ViT, but only small $4\times 4$ patches for most hierarchical models (Fig.~\ref{fig:architecture_overlap}\hyperref[fig:architecture_overlap]{a}).

To address both of these issues, we opt to distinguish tokens from ``mask units''. As described in Fig.~\ref{fig:architecture_overlap}\hyperref[fig:architecture_overlap]{a}, mask units are at the resolution we apply MAE masking, while tokens are the internal resolution of the model (like in \citet{maskfeat,xie2021simmim}). In our case, we mask $32\times32$ pixel regions, meaning one mask unit is $8\times8$ tokens at the start of the network. Once we have made this distinction, we can use a clever trick (Fig.~\ref{fig:architecture_overlap}\hyperref[fig:architecture_overlap]{d}) to evaluate hierarchical models by treating mask units as contiguous, separate from other tokens. Thus, we can continue with our experiments and use MAE with an existing hierarchical vision transformer.

%##################################################################################################
\begin{figure}[t]
\centering
\includegraphics[width=0.99\linewidth]{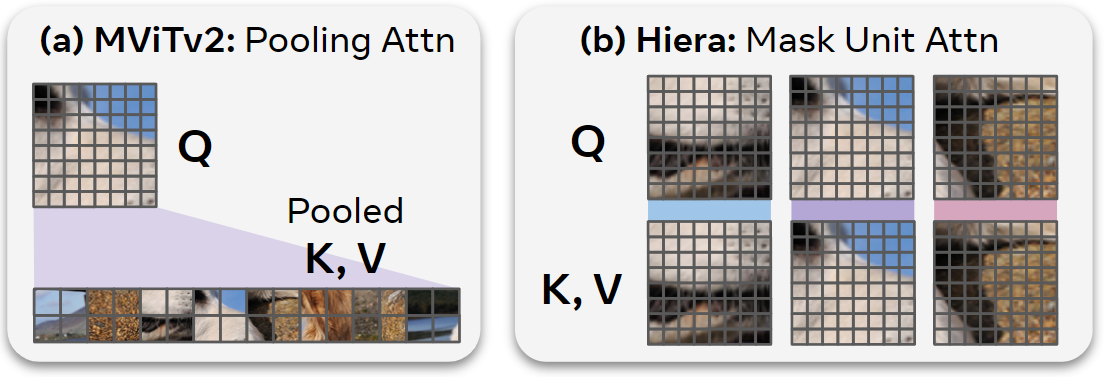}
\vspace{-.3em}
\caption{\textbf{Mask Unit Attention.} MViTv2 uses pooling attention (a) which performs global attention with a pooled version of $K$ and $V$. This can get expensive for large inputs (e.g., for video), so we opt to replace this with ``Mask Unit Attention'' (b) which performs local attention within mask units (Fig.~\ref{fig:architecture_overlap}\hyperref[fig:architecture_overlap]{a}). This has no overhead because we already group tokens into units for masking. We do not have to worry about shifting like in Swin \cite{swin}, because we use global attention in stages 3 and 4 (Fig.~\ref{fig:arch_and_task}).
}
\label{fig:pool_vs_mu_attn}
\vspace{-1em}
\end{figure}
%##################################################################################################

\subsection{Preparing MViTv2} \label{subsec:approach_prep}

We choose MViTv2 as our base architecture, as its small $3\times3$ kernels are affected the least by the separate-and-pad trick described in Fig.~\ref{fig:architecture_overlap}\hyperref[fig:architecture_overlap]{d}, though we likely could have chosen a different transformer and obtained a similar end result. We briefly review MViTv2 below.

\paragraph{MViTv2} \cite{mvitv2} is a \textit{hierarchical model}. That is, it learns multi-scale representations over its four stages. It starts by modeling low level features with a small channel capacity but high spatial resolution, and then in each stage trades channel capacity for spatial resolution to model more complex high-level features in deeper layers.

A key feature of MViTv2 is \textit{pooling attention} (Fig.~\ref{fig:pool_vs_mu_attn}\hyperref[fig:pool_vs_mu_attn]{a}), wherein features are \textit{locally aggregated}---typically using $3\times3$ convolution, before computing self-attention. In pooling attention, $K$ and $V$ are pooled to decrease computation in the first two stages, while $Q$ is pooled to transition from one stage to the next by reducing spatial resolution.
MViTv2 also features \textit{decomposed relative position embeddings} instead of absolute ones and a \textit{residual pooling} connection to skip between pooled $Q$ tokens inside the attention blocks. Note that by default, pooling attention in MViTv2 contain convs with stride 1 even if no downsampling is required.

\paragraph{Applying MAE.}
Since MViTv2 downsamples by $2\times2$ a total of three times (Fig.~\ref{fig:arch_and_task}) and because it uses a token size of $4\times4$ pixels, we employ a mask unit of size $32\times32$. This ensures that each mask unit corresponds to 8$^\text{2}$, 4$^\text{2}$, 2$^\text{2}$, 1$^\text{2}$ tokens in stages 1, 2, 3, 4 respectively, allowing each mask unit to cover at least one distinct token in each stage. Then as described in Fig.~\ref{fig:architecture_overlap}\hyperref[fig:architecture_overlap]{d}, to make sure conv kernels do not bleed into deleted tokens, we shift the mask units to the batch dimension to separate them for pooling (effectively treating each mask unit as an ``image'') and then undo the shift afterward to ensure that self-attention is still global.

\subsection{Simplifying MViTv2} \label{subsec:approach_arch}
In this section we \textit{remove} non-essential components of MViTv2 while training with MAE. In Tab.~\ref{tab:sparsifying_mvit}, we find that we can remove or otherwise simplify \textit{all} of them and still maintain high accuracy for image classification on ImageNet-1K. We use MViTv2-L to ensure our changes work at scale.

\paragraph{Relative Position Embeddings.}
MViTv2 swaps the absolute position embeddings in \citet{vit} for more powerful relative ones added to attention in \textit{each block}. Technically, we could implement a version of this that is compatible with sparse pretraining, but doing so would add a lot of complexity. Instead, we opt to start our study here by undoing this change and using absolute position embeddings instead. As shown in Tab.~\ref{tab:sparsifying_mvit}\hyperref[tab:sparsifying_mvit]{a}, these relative position embeddings are not necessary when training with MAE. Further, absolute position embeddings are much faster.%, especially in video where this change alone almost ?? clip/s.

\paragraph{Removing Convolutions.}
Next, we aim to remove the convs in the model, which are vision specific modules and add potentially unnecessary overhead.
We first attempt to replace every conv layer with maxpools (shown by \citet{mvitv1} to be the next best option), which itself is fairly costly.
The result (Tab.~\ref{tab:sparsifying_mvit}\hyperref[tab:sparsifying_mvit]{b}) drops accuracy by over 1\% on images, but this is to be expected: we've also replaced all of the extra {\scriptsize stride=1} convs with maxpools, which impacts the features significantly (with padding and small mask units, this in effect performs a relu on every feature map). Once we delete those additional {\scriptsize stride=1} maxpools (Tab.~\ref{tab:sparsifying_mvit}\hyperref[tab:sparsifying_mvit]{c}), we nearly return to the accuracy we had before, while speeding up the model by 22\% for images and 27\% for video. At this point, the only pooling layers that remain are for $Q$ at stage transitions and for $KV$ pooling in the first two stages.

%##################################################################################################
\begin{table}[t]
    \begin{center}
        \tablestyle{3pt}{1.05}
        \vspace{-5pt}
        \begin{tabular}{y{115}x{18}x{20}x{1}x{18}x{20}}
        % \toprule
            &\multicolumn{2}{c}{\img{Image}}&&\multicolumn{2}{c}{\vid{Video}}\\
            % \cline{2-3}\cline{5-6}
            Setting & acc. & im/s && acc. & clip/s \\
            \shline
            \gc{MViTv2-L {\scriptsize Supervised}} & \gc{85.3} & \gc{219.8} &\gc{}& \gc{80.5} & \gc{20.5}\\ % 80.5, 8.5
           % \hline
            {\bf \shortname-L} {\scriptsize MAE} \\
            ~a. replace {\scriptsize rel pos} with {\scriptsize absolute} $^*$ & \ul{85.6} & 253.3 && \ul{85.3} & 20.7 \\
            ~b. replace {\scriptsize convs} with {\scriptsize maxpools} $^*$ & 84.4 & $\,$99.9$^\dagger$ && 84.1 & $\,$10.4$^\dagger$ \\
            ~c. delete {\scriptsize stride=1} {\scriptsize maxpools} $^*$   & 85.4 & 309.2 &&  84.3 & 26.2 \\
            ~d. set {\scriptsize kernel size} equal to {\scriptsize stride} & \textbf{85.7} & 369.8 &&  \textbf{85.5} & 29.4 \\
            ~e. delete {\scriptsize q attention residuals} & \ul{85.6} & 374.3 && \textbf{85.5} & 29.8 \\
            \baseline{~f. replace {\scriptsize kv pooling} with {\scriptsize MU attn}} & \baseline{{\ul{85.6}}} & \baseline{\textbf{531.4}} & \baseline{} & \baseline{\textbf{85.5}} & \baseline{\textbf{40.8}} \\
        \end{tabular}
    \end{center}
    \vspace{-1em}
        \caption{\textbf{Simplifying MViTv2.} MViTv2 employs several architectural tweaks to perform well on \gc{supervised training}.
        By progressively removing them in Sec.~\ref{subsec:approach_arch}, we find these bells-and-whistles are \textit{unnecessary} when training with a strong pretext task (MAE). In the process, we create an extremely simple model (Fig.~\ref{fig:arch_and_task}) that is accurate while being significantly faster. We report fp16 inference speed for \img{ImageNet-1K} and \vid{Kinetics-400} on an A100. Our final \shortname{-L} in {\setlength{\fboxsep}{2pt}\colorbox{baselinecolor}{gray}}. $^*$Requires the separate-and-pad trick described in Fig.~\ref{fig:architecture_overlap}\hyperref[fig:architecture_overlap]{d}. $^\dagger$PyTorch's {\scriptsize maxpool3d} interacts unfavorably with this. 
    }
    \label{tab:sparsifying_mvit}
    \vspace{-.4em}
\end{table}
%##################################################################################################

\paragraph{Removing Overlap.}
The remaining maxpool layers still have a kernel size of $3\times3$, necessitating the use of the separate-and-pad trick in Fig.~\ref{fig:architecture_overlap}\hyperref[fig:architecture_overlap]{d} during both training and inference. However, as shown in Fig.~\ref{fig:architecture_overlap}\hyperref[fig:architecture_overlap]{e}, we can avoid this problem entirely if we just do not let these maxpool kernels overlap. That is, if we set the kernel size equal to stride for each maxpool, we can use sparse MAE pretraining \textit{without} the separate-and-pad trick. As shown in Tab.~\ref{tab:sparsifying_mvit}\hyperref[tab:sparsifying_mvit]{d}, this speeds up the model by 20\% on image and 12\% on video while increasing accuracy, likely due to not having to pad.

\paragraph{Removing the Attention Residual.}
MViTv2 adds a residual connection in the attention layer between $Q$ and the output to assist in learning its pooling attention. However, so far we've minimized the number of layers, making attention easier to learn. Thus, we can safely remove it (Tab.~\ref{tab:sparsifying_mvit}\hyperref[tab:sparsifying_mvit]{e}).

%##################################################################################################
\begin{figure}[t]
\centering
\includegraphics[width=0.99\linewidth]{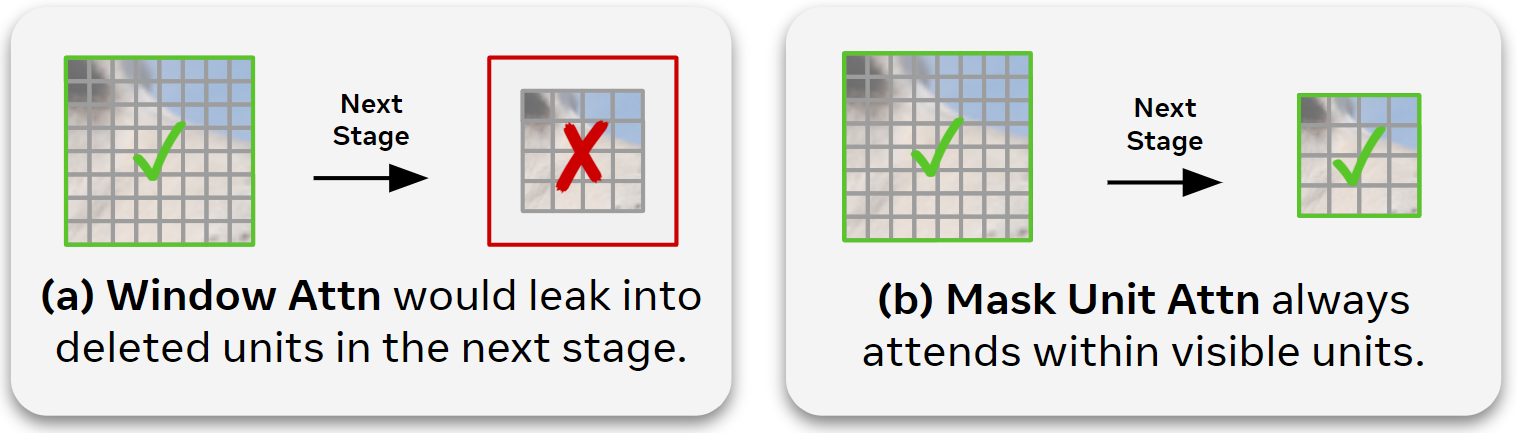}
\vspace{-.5em}
\caption{\textbf{Mask Unit Attn vs. Window Attn.} Window attention (a) performs local attention within a \textit{fixed} size window. Doing so would potentially overlap with deleted tokens during sparse MAE pretraining. In contrast, Mask Unit attention (b) performs local attention within individual mask units, no matter their size.
}
\label{fig:window_vs_mu_attn}
    \vspace{-10pt}
\end{figure}
%##################################################################################################

\paragraph{Mask Unit Attention.}
At this point, the only specialized module left is pooling attention. Pooling $Q$ is necessary to maintain a hierarchical model, but $KV$ pooling is only there to reduce the size of the attention matrix in the first two stages. We can remove this outright, but it would considerably increase the computational cost of the network. Instead, in Tab.~\ref{tab:sparsifying_mvit}\hyperref[tab:sparsifying_mvit]{f} we replace it with an implementationally trivial alternative: local attention within a mask unit.

During MAE pretraining, we already have to separate out mask units at the start of the network (see Fig.~\ref{fig:arch_and_task}). Thus the tokens are already neatly grouped by units once they arrive at attention. We can then simply perform local attention within these units with no overhead. While this ``Mask Unit attention'' is local instead of global like pooling attention (Fig.~\ref{fig:pool_vs_mu_attn}), $K$ and $V$ were only pooled in the first two stages, where global attention isn't as useful. Thus, as shown in Tab.~\ref{tab:sparsifying_mvit}, this change has no impact on accuracy but increases throughput by quite a lot---up to 32\% on video.

Note that mask unit attention is distinct from window attention because it adapts the window size to the size of mask units at the current resolution. Window attention would have a fixed size throughout the network, which would leak into deleted tokens after a downsample (see Fig.~\ref{fig:window_vs_mu_attn}).

\begin{figure}[t]
\centering
\includegraphics[width=\linewidth]{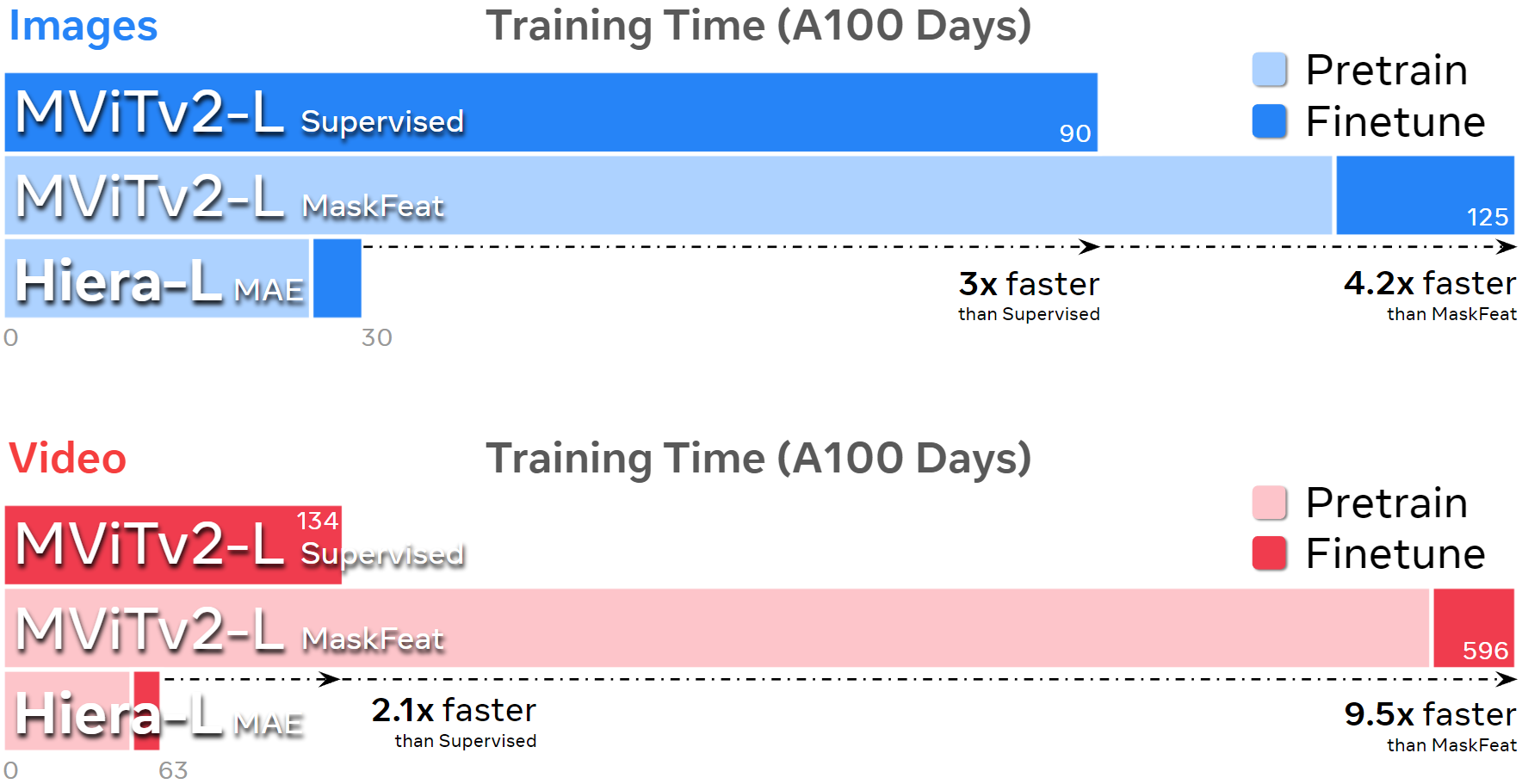}
\vspace{-15pt}
\caption{\textbf{Training time.} Measured in half precision A100 days. Our \shortname{} is \textit{significantly} faster to train than MViTv2 due to being more efficient and benefiting from sparse pretraining (as opposed to MaskFeat). Here, supervised uses 300 epochs for \img{ImageNet-1K} and 200 for \vid{Kinetics-400}, while MaskFeat and MAE use 400 for pretraining on \img{images} and 800 on \vid{video} followed by 50 epochs of finetuning for both. Note that \shortname{-L} at 200 epochs of pretraining (81.8) already outperforms MViTv2-L supervised (80.5) on \vid{video}, making it $5.6\times$ faster to obtain higher accuracy. }
\label{fig:sparsifying_mvit_train_speed}
\vspace{-0.05in}
\end{figure}

\paragraph{\name{}.}
The result of these changes is an extremely simple and efficient model, which we denote ``\name{}''. \shortname{} is \textit{2.4$\times$ faster} on images and \textit{5.1$\times$ faster} on video than the MViTv2 we started with and is actually \textit{more accurate} because of MAE. Furthermore, because \shortname{} supports sparse pretraining, the results in Tab.~\ref{tab:sparsifying_mvit} are extremely fast to obtain. In fact, to obtain superior accuracy on images, \shortname{-L} is \textit{3$\times$ faster} to train than a supervised MViTv2-L (Fig.~\ref{fig:sparsifying_mvit_train_speed}). For video, \citet{maskfeat} report 80.5\% using a \textit{cut down} version of MViTv2 with double the $KV$ stride in the first 3 stages. Compared to this model, our \shortname-L obtains 85.5\% in 800 pretrain epochs while being \textit{2.1$\times$ faster} to train (Fig.~\ref{fig:sparsifying_mvit_train_speed}).
All benchmarks in this paper are on an A100 with fp16 (as this setting is most useful in practice) unless noted otherwise.

While we used \shortname{-L} for the experiments in this section, we can of course instantiate it in different sizes, e.g. Tab.~\ref{tab:hiera_configs}.

\begin{table}[h]
    \centering
    \resizebox{\linewidth}{!}{
    \tablestyle{1pt}{1.1}
        \begin{tabular}{lcccrr}
         model                          & \#Channels       &  \#Blocks      & \#Heads & FLOPs & Param \\
        \shline
        \shortname{-T}                  & {\footnotesize [96-192-384-768]} & {\footnotesize [1-2-7-2] }    & {\footnotesize [1-2-4-8]} & 5G      & 28M \\
        \shortname{-S}                  & {\footnotesize [96-192-384-768]} & {\footnotesize [1-2-11-2] }    & {\footnotesize [1-2-4-8]} & 6G      & 35M \\
        \shortname{-B}                  & {\footnotesize [96-192-384-768]} & {\footnotesize [2-3-16-3] }    & {\footnotesize [1-2-4-8]} & 9G      & 52M \\
        \shortname{-B+}                 & {\footnotesize [112-224-448-896]} & {\footnotesize [2-3-16-3] }    & {\footnotesize [2-4-8-16]} & 13G      & 70M \\
        \shortname{-L}                  & {\footnotesize [144-288-576-1152]} &{\footnotesize  [2-6-36-4] }    & {\footnotesize [2-4-8-16]} & 40G       & 214M \\
        \shortname{-H}                  & {\footnotesize [256-512-1024-2048]} & {\footnotesize [2-6-36-4] }    & {\footnotesize [4-8-16-32]} & 125G       & 673M \\
        \end{tabular}% }
    }
    \vspace{-10pt}
    \caption{\textbf{Configuration for \name{} variants}. \#Channels, \#Blocks and \#Heads specify the channel width, number of \shortname blocks
and heads in each block for the four stages, respectively. FLOPs are measured for image classification with 224 $\times$ 224 input. The
stage resolutions are [56$^2$, 28$^2$, 14$^2$, 7$^2$]. We introduce B+ for more direct comparison against prior work with slower B models.
    }
    \label{tab:hiera_configs}
\end{table}

%##################################################################################################
%##################################################################################################

\section{MAE Ablations}
\label{sec:mae_ablations}
In this section, we ablate MAE pretraining settings in \shortname{} for both images and video, using ImageNet-1K (IN1K, \citet{imagenet}) and Kinetics-400 (K400, \citet{k400}). Like in \citet{mae,mae-st}, we ablate using our large model, \shortname{-L}, to ensure that our method works at scale. We evaluate performance by finetuning. All metrics are top-1 accuracies using standard evaluation protocols---a single (resized) center crop on IN1K and 3 spatial $\times$ 5 temporal views on K400.

%##################################################################################################
% overall table of all ablations
\begin{table*}[t]
\vspace{-10pt}
\centering
%#################################################
%#################################################
\subfloat[
\textbf{Multi-Scale Decoder.} \shortname{} being \textit{hierarchical}, using multi-scale information for decoding brings significant gains.
\label{tab:ablation_decoder_multiscale}
]{
\centering
\begin{minipage}{0.29\linewidth}{\begin{center}
\tablestyle{4pt}{1.05}
\begin{tabular}{x{40}x{20}x{20}}
 multi-scale & \img{image} &  \vid{video}  \\
\shline
\xmark & 85.0 & 83.8 \\
\chkmark & \baseline{\textbf{85.6}} & \baseline{\textbf{85.5}}\\
\\
\end{tabular}
\end{center}}\end{minipage}
}
\hspace{1.2em}
%#################################################
%#################################################
\subfloat[
\textbf{Mask ratio.} High masking ratios lead to good performance, with video benefiting from higher masking than image modality.
\label{tab:ablation_mask_ratio}
]{
\centering
\begin{minipage}{0.33\linewidth}{\begin{center}
\tablestyle{4pt}{1.05}
\begin{tabular}{x{20}x{22}|x{20}x{20}}
mask & \img{image} & mask &  \vid{video} \\
\shline
0.5 &  85.5 & 0.75 & 84.9\\
0.6 & \baseline{\textbf{85.6}} & 0.9 & \baseline{\textbf{85.5}} \\
0.7 & 85.3  & 0.95 & 84.4\\
\end{tabular}
\end{center}}\end{minipage}
}
\hspace{1.2em}
%#################################################
%#################################################
\subfloat[
\textbf{Reconstruction target.} Both pixel and HOG targets result in strong performance.
\label{tab:ablation_target}
]{
\centering
\begin{minipage}{0.29\linewidth}{\begin{center}
\tablestyle{4pt}{1.05}
\begin{tabular}{x{20}x{20}x{20}}
 target & \img{image} &  \vid{video}  \\
\shline
pixel & \baseline{85.6} & \baseline{85.5}\\
HOG & \textbf{85.7} & \textbf{86.1} \\
\\
\end{tabular}
\end{center}}\end{minipage}
}
\hspace{1em}
\vspace{-10pt}
%#################################################
%#################################################
\subfloat[
\textbf{Drop path rate.} Surprisingly, we find drop path important during MAE pretraining, especially for video, unlike in \citet{mae, mae-st}.  \label{tab:ablation_dpr}
]{
\centering
\begin{minipage}{0.29\linewidth}{\begin{center}
\tablestyle{4pt}{1.05}
\begin{tabular}{x{18}x{20}x{20}}
dpr & \img{image} & \vid{video}  \\
\shline
0.0 & 85.2 & 84.5\\
0.1 & 85.6 & 85.4 \\
0.2 & \baseline{\textbf{85.6}} & \baseline{\textbf{85.5}}\\
0.3 & 85.5 & 85.2\\
\end{tabular}
\end{center}}\end{minipage}
}
\hspace{1.2em}
%#################################################
%#################################################
\subfloat[
\textbf{Decoder depth.} We find that a deeper decoder than in \citet{mae-st} works better for video.
\label{tab:ablation_decoder_depth}
]{
\centering
\begin{minipage}{0.29\linewidth}{\begin{center}
\tablestyle{4pt}{1.05}
\begin{tabular}{x{24}x{20}x{20}}
 depth & \img{image} &  \vid{video}  \\
\shline
4 & 85.5 & 84.8\\
8 & \baseline{\textbf{85.6}} & \baseline{\textbf{85.5}}\\
12 & 85.5 & 85.4\\
\\
\end{tabular}
\end{center}}\end{minipage}
}
\hspace{1.2em}
%#################################################
%#################################################
\subfloat[
\textbf{Pretraining schedule.} Our pretraining follows the same trend as \citet{mae}, benefiting significantly from longer training.  \label{tab:ablation_pretraining_schedule}
]{
\centering
\begin{minipage}{0.29\linewidth}{\begin{center}
\tablestyle{4pt}{1.05}
\begin{tabular}{x{24}x{20}x{20}}
epochs & \img{image} & \vid{video} \\
\shline
400 & \baseline{85.6} & 84.0 \\
800 & 85.8 & \baseline{85.5} \\
1600 & \textbf{86.1} & 86.4\\
3200 & \textbf{86.1} & \textbf{87.3}\\
\end{tabular}
\end{center}}\end{minipage}
}
\vspace{1.0em}
\caption{\textbf{Ablating MAE pretraining with \name-L.} For each ablation, we use \img{400} (\vid{800}) epochs of sparse MAE pretraining for \img{IN1K} (\vid{K400}) and 50 epochs of dense finetuning unless otherwise noted. Our default$^\dagger$ settings are marked in \colorbox{baselinecolor}{gray}. For design choices not ablated here, we find the defaults in \cite{mae,mae-st} to be appropriate.  {\tiny $\dagger$} {\small default pretraining length for the rest of the paper is \img{1600} (\vid{3200}) epochs, unless otherwise noted.}
}
\label{tab:ablations} \vspace{-.7em}
\end{table*}
%##################################################################################################

% MAE pretraining
\paragraph{Multi-Scale decoder.}
While \citet{mae,mae-st} use the tokens from the \textit{last} block of the encoder as the input to the decoder, \shortname{} being \textit{hierarchical} permits more flexibility: as in \citet{gao2022mcmae}, we can use \textit{multi-scale} information by fusing representations from all stages, which brings large gains in both modalities (Tab.~\ref{tab:ablation_decoder_multiscale}).

\paragraph{Masking ratio.}
\citet{mae-st} find video to require a much higher masking ratio than images, suggesting higher information redundancy.
We observe a similar trend in Tab.~\ref{tab:ablation_mask_ratio} with optimal masking ratios of 0.6 for images but 0.9 for video.
Our optimal ratio for images, 0.6, is slightly lower than the 0.75 used in \citet{mae}. We expect this is due to increased difficulty of the pretext task from using a 32 $\times$ 32 mask unit instead of 16 $\times$ 16 as in \citet{mae}.
Interestingly, we find the same 0.9 masking ratio to be appropriate for video as in \citet{mae-st}.
This could be because they actually find 0.95 to work optimally if allowed to train twice as long. With our increased task difficulty, 0.9 works out to be best.

\paragraph{Reconstruction target.} We find (Tab.~\ref{tab:ablation_target}) that both pixel (w\slash~norm) and HOG \cite{hog} targets result in strong performance. While HOG targets results in slightly better performance for the default number of pretraining epochs we use in ablations, we found that with longer training HOG targets achieve the same performance as pixel targets for video, but slightly worse for images.

\paragraph{Droppath rate.}
The original MAE pretraining recipes \cite{mae,mae-st} explicitly do not use drop path \cite{droppath} during pretraining, instead opting to only do so while finetuning. However, our \shortname{-L} has twice the depth of a ViT-L model: 48 for \shortname{-L} \textit{vs.} 24 for ViT-L. While each layer individually has a lower parameter count, due to the sheer depth of \shortname{}, there could be a significant benefit from drop path.

In Tab.~\ref{tab:ablation_dpr}, we ablate applying drop path during pretraining (finetuning employs drop path by default) and find significant gains. This is surprising because it means that without drop path, \name{} can overfit to the MAE task.

\paragraph{Decoder depth.}
We find a significant benefit from a deeper decoder than previous work use for video \cite{mae-st}, see Tab.~\ref{tab:ablation_decoder_depth}. This brings the decoder for video in line with images \cite{mae}.

\paragraph{Pretraining schedule.}
Several masked image modeling approaches \cite{mae,maskfeat} have found benefits from longer pretraining schedules, often using up to 1600 epochs. In Tab.~\ref{tab:ablation_pretraining_schedule}, we observe the same trend for \shortname{}, increasing +0.5\% over 400 epochs on IN1K. In fact, \shortname{}'s accuracy at 400ep is +0.7\% higher than ViT-L MAE (84.9\%) at the same number of epochs but only +0.2\% higher at 1600 epochs---suggesting that \shortname{} is a more \textit{efficient} learner. On K400, even with only 800 epochs of pretraining, \shortname{} outperforms the previous SotA result that uses 1600 epochs (85.2\%). Gains from longer training saturate less quickly on video, with a large 0.9\% gain from 800 epochs to 1600 epochs and beyond.

\section{Video Results} \label{sec:video_experiments}
We report our results on video recognition. All models input 16 frames of $224^2$ pixels unless otherwise specified. For video, mask units are $2\times32\times32$ px (i.e., $1\times8\times8$ tokens as before). The rest of the model is the same as for images.

\begin{table}[t]
    \centering
    \resizebox{\linewidth}{!}{
    \tablestyle{7pt}{1.1}
        \begin{tabular}{llcrr}
        backbone                        & pretrain      &  acc.               & FLOPs (G)                            & Param \\
        \shline        
        ViT-B                           &MAE             & 81.5                & \flops{180}{3}{5}                & 87M \\  % VideoMAE
        \ours{\shortname{-B}}           &MAE     & \ul{84.0}      & \flops{\textbf{102}}{3}{5}       & \textbf{51M} \\
        \ours{\shortname{-B+}}           &MAE     & \textbf{85.0}      & \flops{\ul{133}}{3}{5}       & \ul{69M} \\
        \shline
        MViTv2-L                        &-               & 80.5                & \flops{\textbf{377}}{1}{10}               & \ul{218M} \\
        MViTv2-L                        &MaskFeat        & 84.3                & \flops{\textbf{377}}{1}{10}               & \ul{218M} \\
        ViT-L                           &MAE             & \ul{85.2}                & \flops{597}{3}{5}                & 305M \\  % VideoMAE
        \ours{\shortname{-L}}           &MAE        & \textbf{87.3}                & \flops{\ul{413}}{3}{5}       & \textbf{213M} \\
        \shline
        ViT-H                           &MAE     & 86.6                & \flops{1192}{3}{5}                & \textbf{633M}  \\  % VideoMAE
        \ours{\shortname{-H}}           &MAE     & \textbf{87.8}       & \flops{\textbf{1159}}{3}{5}                & 672M\\
        \end{tabular}
    }
    \caption{\textbf{K400 results}. \shortname{} improves on previous SotA by a large amount, while being lighter and faster. FLOPs are reported as inference FLOPs $\times$ spatial crops $\times$ temporal clips.
    }
    \label{tab:sota:k400}
    % \vspace{-0.5cm}
\end{table}

\paragraph{Kinetics-400,-600,-700.} In Tab.~\ref{tab:sota:k400}, we compare \shortname{} trained with MAE to the SotA on Kinetics-400 \cite{k400} at a system level. We compare to MViTv2-L~\cite{mvitv2} pretrained with MaskFeat~\cite{maskfeat} and ViT~\cite{vit} pretrained with MAE on video~\cite{mae-st,tong2022videomae}. \shortname{-L} brings large gains (+\textbf{2.1}\%) over previous SotA \cite{mae-st, tong2022videomae}, while using {\tiny$\sim$}45\% \textit{fewer} flops, being {\tiny$\sim$}43\% \textit{smaller} and \textbf{2.3}$\times$ faster (Fig.~\ref{fig:imnet1k_inference_speed}). In fact, \shortname{-L} \textit{significantly outperforms} (+\textbf{0.7}\%) models \textit{one tier higher}, while being \textbf{3$\times$} \textit{smaller} and \textbf{3.5$\times$} \textit{faster}. \shortname{-L} achieves a gain of +\textbf{6.8}\% over the corresponding MViTv2-L supervised baseline. Going one tier up in size, \shortname{-H} improves performance over previous SotA by +\textbf{1.2}\%, \textit{establishing a new SotA} for $224^2$ without external data. We show similarly large improvements over the art on K600 (+\textbf{1.9}\%) and K700 (+\textbf{2.8}\%) in Tab.~\ref{tab:sota:k600,700}, with our H models bringing even further gains.

\begin{table}[t]
    \centering
    \vspace{-0.1in}
    \resizebox{\linewidth}{!}{
        \subfloat[{\small \textbf{Kinetics-600}} video classification\label{tab:k600}]{
        \begin{tabular}{llcrr}
        backbone        & pretrain      &  acc.               & FLOPs (G)                            & Param\\
        \shline
        \gc{MViTv2-L}        &\gc{Sup, IN-21K}     &\gc{85.8}                & \gc{\flops{377}{1}{10}}               & \gc{218M} \\
        MViTv2-L        &MaskFeat        & \ul{86.4}                & \flops{\textbf{377}}{1}{10}               & \ul{218M} \\
        \ours{\shortname{-L}}           &MAE     & \textbf{88.3}                & \flops{\ul{413}}{3}{5}       & \textbf{213M} \\
        \shline
        \ours{\shortname{-H}}           &MAE     & \textbf{88.8}       & \flops{1159}{3}{5}                & 672M\\
        \end{tabular}% }
        }
        }
        \resizebox{\linewidth}{!}{
        \hfill
        \subfloat[{\small \textbf{Kinetics-700}} video classification\label{tab:k700}]{
        \begin{tabular}{llcrr}
        backbone        & pretrain      &  acc.               & FLOPs (G)                            & Param \\
        \shline
        \gc{MViTv2-L}        &\gc{Sup, IN-21K}     &\gc{76.7}                & \gc{\flops{377}{1}{10}}               & \gc{218M} \\
        MViTv2-L        &MaskFeat        & \ul{77.5}                & \flops{\textbf{377}}{1}{10}               & \ul{218M} \\
        \ours{\shortname{-L}}           &MAE     & \textbf{80.3}                & \flops{\ul{413}}{3}{5}       & \textbf{213M} \\
        \shline
        \ours{\shortname{-H}}           &MAE     & \textbf{81.1}       & \flops{1159}{3}{5}                & 672M\\
        \end{tabular}% }
        }
    }
    \vspace{-0.1in}
    \caption{\textbf{K600 and K700 results}. \shortname{} improves over SotA by a large margin. FLOPs reported as inference FLOPs $\times$ spatial crops $\times$ temporal clips. Approaches using extra data are \gc{de-emphasized}.
    }
    \label{tab:sota:k600,700}
    \vspace{-0.1in}
\end{table}

\paragraph{Something-Something-v2 (SSv2).} In Tab.~\ref{tab:sota:ssv2}, we compare our \shortname{} with the current art on SSv2~\cite{goyal2017something} at a system level: MViTv2-L~\cite{mvitv2} pretrained with MaskFeat~\cite{maskfeat} and ViT~\cite{vit} pretrained with MAE on video~\cite{tong2022videomae}. When pretrained on K400, \shortname{-L} outperforms the runner-up method MaskFeat by +$\textbf{0.6}\%$, but \shortname{} is \textit{dramatically} more efficient, using 16 frames at $224^2$ resolution \textit{vs}. 40 frames at $312^2$ resolution in MaskFeat, effectively using ~\textbf{3.4$\times$} \textit{fewer} FLOPs. When pretrained on SSv2, \shortname{-L} achieves $\textbf{75.1}\%$, outperforming ViT-L pretrained with MAE, by +\textbf{0.8}\%, while using {\tiny$\sim$}45\% \textit{fewer} flops and being {\tiny$\sim$}43\% \textit{smaller}. Our \shortname{-L}$_{32}$ model further achieves  $\textbf{76.5}\%$, SotA among approaches trained only on SSv2.

\begin{table}[t!]
    \centering
    \resizebox{\linewidth}{!}{
    \tablestyle{4pt}{1.1}
        \begin{tabular}{llcrr}
        backbone & pretrain &   acc.  & FLOPs (G)  & Param \\
        \shline
        \multicolumn{3}{@{}l}{\emph{K400 pretrain}} \\
        \hline
         ViT-L & supervised  & 55.7 & 598$\times$3$\times$1 & 304M \\ 
         MViTv2-L$_{40,312}$ & MaskFeat   & 74.4 & 2828$\times$3$\times$1 & \ul{218M}\\
         ViT-L & MAE  &  74.0 & \ul{597}$\times$3$\times$2 & 305M\\ 
         \baseline{\shortname{-L}} & MAE  & \ul{74.7} & {\bf 413}$\times$3$\times$1 & \bf 213M \\ 
         \baseline{\shortname{-L}} & MAE  & \bf{75.0} & {\bf 413}$\times$3$\times$2 &  \bf 213M \\
         \midrule
         \multicolumn{3}{@{}l}{\emph{SSv2 pretrain}} \\
         \hline
         ViT-L & MAE  & 74.3 & 597$\times$3$\times$2 & 305M\\ 
         \baseline{\shortname{-L}} & MAE  & \ul{74.9} & {\bf 413}$\times$3$\times$1 & \bf{213M}\\
         \baseline{\shortname{-L}} & MAE  & \bf 75.1 & {\bf 413}$\times$3$\times$2 & \bf{213M}\\ 
         \hline
         ViT-L$_{32}$ & MAE  & 75.4 & 1436$\times$3$\times$1 & 305M\\
         \baseline{\shortname{-L}$_{32}$} & MAE  & \bf{76.5} & \textbf{1029}$\times$3$\times$1 & \bf{213M}\\ 
        \end{tabular}
    }
    \caption{\textbf{SSv2 results} pretrained on Kinetics-400 and SSv2. \shortname{} improves over SotA by a large margin. We report inference FLOPs $\times$ spatial crops $\times$ temporal clips.}
    \label{tab:sota:ssv2}
\end{table}

\paragraph{Transferring to action detection (AVA).} We evaluate transfer learning of K400/K600/K700 pretrained \shortname{} on action detection using AVA v2.2 dataset~\cite{gu2018ava}. In Tab.~\ref{tab:sota:ava} we compare the pretrained \shortname{} with SotA methods, MViTv2 with MaskFeat~\cite{maskfeat} and ViT with MAE on video~\cite{tong2022videomae,mae-st} at system level, and report mean average precision (mAP). Our K400 pretrained \shortname{}-L outperforms an MAE pretrained ViT-L by +\textbf{2.8}\% and an MViTv2-L$_{40,312}$ MaskFeat by +\textbf{1.3}\% mAP while \shortname{}-L has fewer FLOPs and parameters. Our \shortname{}-H outperforms an MAE pretrained ViT-H by +\textbf{3.0}\% mAP. We observe similar performance improvement of the K600/K700 pretrained \shortname{} as well. Specifically, the K700 pretrained \shortname{}-H outperforms an MAE pretrained ViT-H by +\textbf{3.2}, establishing a new SotA.

\begin{table}[t!]
    \centering
    \resizebox{\linewidth}{!}{
    \tablestyle{4pt}{1.1}
        \begin{tabular}{llcrr}
        backbone & pretrain &   mAP  & FLOPs (G)  & Param \\
        \shline
        \multicolumn{3}{@{}l}{\emph{K400 pretrain}} \\
        \hline
         ViT-L & supervised & 22.2 & 598 & 304M\\
         MViTv2-L$_{40,312}$ & MaskFeat   & \ul{38.5} & 2828 & \ul{218M}\\
         
         ViT-L & MAE &  37.0 & \ul{597} & 305M\\ 
         \baseline{\shortname{-L}} & MAE
         & \bf 39.8 & {\bf 413} & \bf 213M \\
         
         \hline
         ViT-H & MAE &  39.5 & 1192 & \bf{633M}\\ 
         \baseline{\shortname{-H}} & MAE  &  {\bf 42.5} & \bf 1158 & 672M \\
   
        \midrule
        \multicolumn{3}{@{}l}{\emph{K600 pretrain}} \\
        \hline
        ViT-L & MAE &  38.4 & \ul{598} & 304M\\ 
        MViTv2-L$_{40,312}$ & MaskFeat   & \ul{39.8} & 2828 & \ul{218M}\\
        \baseline{\shortname{-L}} & MAE & \bf 40.7 & {\bf 413} & \bf 213M \\
        \hline
        ViT-H & MAE &  40.3 & 1193 & \bf 632M\\ 
         
        \baseline{\shortname{-H}} & MAE  &  {\bf 42.8} & \bf 1158 & 672M \\
        \midrule
        \multicolumn{3}{@{}l}{\emph{K700 pretrain}} \\
        \hline
        ViT-L & MAE &  39.5 & 598 & 304M\\ 
        \baseline{\shortname{-L}} & MAE & \bf 41.7 & {\bf 413} & \bf 213M \\
        \hline
        ViT-H & MAE &  40.1 & 1193 & \bf 632M\\ 
        \baseline{\shortname{-H}} & MAE  &  {\bf 43.3} & \bf 1158 & 672M \\

        \end{tabular}
    }
    \vspace{-0.1in}
    \caption{\textbf{AVA v2.2 results} pretrained on Kinetics. \shortname{} improves over SotA by a large margin. All inference FLOPs reported with a center crop strategy following \citet{mvitv1}. }
    \label{tab:sota:ava}
    \vspace{-10pt}
\end{table}

\section{Image Results} \label{sec:image_experiments}
We first evaluate performance on IN1K and then transfer to other image recognition, detection, and segmentation tasks.

\subsection{Performance on ImageNet-1K}

In Tab.~\ref{tab:sota:in1k}, we perform a system-level comparison of \shortname{} trained with MAE to relevant prior work. First, we observe that the supervised MViTv2 baselines are already quite strong, with MViTv2-B (L) reaching 84.4 (85.3) top-1 accuracy---better than several approaches that use pretraining (e.g. ViT-B MAE). This showcases the significant benefits that convolutions give in the supervised setting, \textit{especially} at the \textit{base} model size and \textit{lower}. Remarkably, even at this size, \shortname{-B} \textit{without} using any bells-and-whistles (e.g., convs), is able to reach \textbf{84.5}\% (slightly) outperforming MViTv2-B; MCMAE-B achieves a higher accuracy, but the model is significantly heavier. Our \shortname{-B+} model handily outperforms it in both speed (Fig.~\ref{fig:imnet1k_inference_speed}) and accuracy. Going \textit{even smaller}, \shortname{-S}, -T demonstrate remarkably strong performance - in a scale regime where convolutions have historically dominated, consistent with our core premise that good spatial biases can be \textit{learned}.

At our default scale, \shortname{-L} MAE reaches an accuracy of \textbf{86.1}\%, a significant +\textbf{0.8}\% gain over MViTv2-L; it also (slightly) outperforms ViT-L MAE, which is 42\% larger and has $1.6\times$ the FLOPs, by +\textbf{0.2}\%. 
Note that while we adopted the MAE pretraining in this work due to its efficient sparse pretraining, \shortname{-L} is readily compatible with complementary, \textit{orthogonal} approaches, e.g. using an EMA teacher \cite{splitmask,data2vec}.

\begin{table}[t]
    \centering
    \resizebox{\linewidth}{!}{
    \tablestyle{7pt}{1.1}
        \begin{tabular}{llccr}
        backbone                       & pretrain                           & acc.    & FLOPs (G)  & Param \\
        \shline
         Swin-T                         &                                   & 81.3      &\textbf{5}     &29M\\ % from swin;
         MViTv2-T                       &                                   & \ul{82.3}      &\textbf{5}     &\textbf{24M}\\ % from MViTv2
        \hline
        \baseline{\shortname{-T}}       & MAE                               & \textbf{82.8}    &\textbf{5}     &\ul{28M}\\
        \shline
         Swin-S                         &                                   & 83.0      &9     &\ul{50M}\\ % from swin;
         MViTv2-S                       &                                   & \ul{83.6}      &\ul{7}     &\textbf{35M}\\ % from MViTv2
         \hline
        \baseline{\shortname{-S}}       & MAE                               & \textbf{83.8}    &\textbf{6}     &\textbf{35M}\\
        \shline
         ViT-B                          &                                   & 82.3      &18     &87M\\ % MAE-recipe; 
         Swin-B                         &                                   & 83.3      &15     &88M\\ % from swin;
         MViTv2-B                       &                                   & 84.4      &\ul{10}     &\textbf{52M}\\ % from MViTv2
         \hline
         \gc{ViT-B}                     & \gc{BEiT}, \gc{\tiny{DALLE}}      &\gc{83.2} &\gc{18}     &\gc{87M}\\
         ViT-B                          & MAE                               & 83.6    &18     &87M\\ 
         ViT-B                          & MaskFeat                          & 84.0    &18     &87M\\ 
         Swin-B                         & SimMIM                            & 83.8    &15     &88M\\ % from SimMIM
         MCMAE-B                        & MCMAE                             & \ul{85.0}   &28      &88M\\ % from MCMAE
        \baseline{\shortname{-B}}       & MAE                               & 84.5    &\textbf{9}     &\textbf{52M}\\
        \baseline{\shortname{-B+}}      & MAE                               & \textbf{85.2}    &13     &\ul{70M}\\
        \shline
         ViT-L                          &                                  & 82.6   &62    & 304M \\ % from MAE
         MViTv2-L                       &                                  & 85.3   &42    & 218M \\ % from MViTv2
         \hline
         \gc{ViT-L}                     & \gc{BEiT}, \gc{\tiny{DALLE}}     & \gc{85.2} &62     &304M\\
         ViT-L                          & MAE                              & 85.9      &62     &304M\\
         ViT-L                          & MaskFeat                         & 85.7      &62     &304M\\
         Swin-L                          & SimMIM                           & 85.4      &\textbf{36}     &\textbf{197M}\\
         MCMAE-L                         & MCMAE                            & \textbf{86.2}      &94     &323M\\
         \baseline{\shortname{-L}}      & MAE                              & \ul{86.1}          &\ul{40}     &\ul{214M}\\
         \shline
         ViT-H                          &                                  & \ul{83.1} & \ul{167} & \textbf{632M}\\
         \hline
         ViT-H                          & MAE                              & \textbf{86.9}         & \ul{167} & \textbf{632M}\\
         \baseline{\shortname{-H}}      & MAE                              & \textbf{86.9}  & \textbf{125}   & \ul{673M} \\
        \end{tabular}
    }
    \vspace{-0.1in}
    \caption{\textbf{ImageNet-1K} comparison to previous MIM approaches. We \gc{de-emphasize} approaches using extra data and indicate the source of extra data.
    }
    \label{tab:sota:in1k}
    \vspace{-10pt}
\end{table}

\subsection{Transfer learning experiments}

Here, we perform transfer learning experiments on downstream classification, detection, and segmentation tasks.

\paragraph{Classification on iNaturalists and Places.} In Tab.~\ref{tab:cls_transfer} we evaluate transfer learning performance on downstream iNaturalist~\cite{van2018inaturalist} and Places~\cite{NIPS2014_3fe94a00} datasets. We finetune the ImageNet-1K pretrained \shortname{} on iNaturalist 2017, 2018, and 2019, and Places 365. \shortname{} consistently outperforms ViT pretrained with MAE \cite{mae}, indicating that our \shortname{}-L and \shortname{}-H architectures are effective outside of just ImageNet.

%##################################################################################################
\begin{table}[t]
\tablestyle{5pt}{1.05}
\begin{tabular}{l x{36}x{36}x{36}x{36}}
\iffalse
{Dataset}
& ViT-L & \baseline{\shortname-L} & ViT-H & \baseline{\shortname-H} \\
\shline
iNat 2017 & 75.7 & \bf 76.8 & 79.3 & \bf 79.6\\
iNat 2018 & 80.1 & \bf 80.9 & 83.0 & 81.7\\
iNat 2019 & 83.4 & \bf 84.3 & \textbf{85.7} & \textbf{85.7}\\
Places365 & 59.4 & \bf 59.5  & \bf 59.8 & \bf 59.8\\
\fi
{backbone} & iNat17 & iNat18 & iNat19 & Places365 \\
\shline
ViT-B & 70.5 & 75.4 & 80.5 & 57.9\\
\baseline{\shortname{-B}} & \ul{73.3} & \ul{77.9} & \ul{83.0} & \ul{58.9}\\
\baseline{\shortname{-B+}} & \bf 74.7 & \bf 79.9 & \bf 83.1 & \bf 59.2 \\
\hline
ViT-L & 75.7 & 80.1 & 83.4 & 59.4\\
\baseline{\shortname{-L}} & \bf 76.8 & \bf 80.9 & \bf 84.3 & \bf 59.6\\
\hline
ViT-H & 79.3 & 83.0 & \bf 85.7 & 59.8\\
\baseline{\shortname{-H}} & \bf 79.6 & \bf 83.5 & \bf 85.7 & \bf 60.0\\
\hline
ViT-H$_{448}$ & 83.4 & 86.8 & 88.3 & 60.3\\
\baseline{\shortname{-H}$_{448}$} & \bf 83.8 & \bf 87.3 & \bf 88.5 & \bf 60.6\\

\end{tabular}
\vspace{-0.1in}
\caption{\textbf{Transfer learning} on iNaturalists and Places datasets.
}
\vspace{-10pt}
\label{tab:cls_transfer}
\end{table}
%##################################################################################################

\paragraph{Object detection and segmentation on COCO.} We finetune Mask R-CNN~\cite{maskrcnn} with different pretrained backbones on the COCO dataset~\cite{coco}. We report \boxAP and \maskAP~for object detection and instance segmentation. We utilize the training recipe following ViTDet~\cite{vitdet} and incorporate multi-scale features from \shortname{} with a Feature Pyramid Network (FPN, \citet{fpn}) as described in the original paper.

In Tab.~\ref{tab:coco}, our \shortname~with MAE pretraining demonstrates a strong scaling behavior when compared models with supervised pretraining such as MViTv2~\cite{mvitv2}, while being consistently faster. For example, \shortname-L is \textbf{+1.8} \boxAP~higher than MViTv2-L (55.0 \emph{vs.} 53.2) with a \textbf{24}\% reduction in inference time. Even when compared to MViTv2 using ImageNet-21K pretraining, \shortname-L still performs \textbf{+1.4} \boxAP better than MViTv2-L.

When compared to the state-of-the-art method, ViTDet, our \shortname{} models achieve comparable results while having faster inference and a lower operation count. For example, \shortname-B shows +0.6 higher \boxAP than ViTDet-B with \textbf{34}\% fewer parameters and \textbf{15}\% lower inference time. Additionally, \shortname-B+ achieves \textbf{+1.9} boxAP improvements while having lower inference time and model complexity \textit{vs}. ViTDet-B.  For the large model, \shortname-L is consistently faster than ViTDet-L with only a slightly lower accuracy.

%##################################################################################################

\begin{table}[t]
    \centering
    \resizebox{\linewidth}{!}{
    \tablestyle{5pt}{1.05}
   \begin{tabular}{@{}y{42}y{36}|x{20}x{20}|x{20}x{20}x{20}}
        backbone &  pretrain &
        {\boxAP} & {\maskAP} & FLOPs & params & time\\
        \shline
        \hline
        \gc{Swin-B} & \gc{Sup, \scriptsize{21K}} & \gc{51.4} & \gc{45.4} & \gc{0.7T}  & \gc{109M} & \gc{164\scriptsize{ms}}\\
        \gc{MViTv2-B} & \gc{Sup, \scriptsize{21K}} & \gc{53.1} & \gc{47.4} & \gc{0.6T} & \gc{73M} & \gc{208\scriptsize{ms}}\\ 
        Swin-B & Sup & 50.1 & 44.5 & 0.7T  & 109M & \textbf{164\scriptsize{ms}}\\
        MViTv2-B & Sup & \ul{52.4} & \ul{46.7} & \textbf{0.6T} & \textbf{73M} & 208\scriptsize{ms}\\ 
        ViTDet-B & MAE & 51.6 & 45.9 & 0.8T & 111M & 201\scriptsize{ms}\\
        \baseline{\shortname-B} & {MAE} & {52.2} & {46.3} & \textbf{{0.6T}} & \textbf{{73M}}  & \ul{173\scriptsize{ms}}\\
        \baseline{\shortname-B+} & {MAE} & \textbf{53.5} & \textbf{47.3} & {\textbf{0.6T}} & {{92M}}  & {192\scriptsize{ms}}\\
        \hline
        \gc{Swin-L} & \gc{Sup, \scriptsize{21K}} & \gc{52.4} & \gc{46.2} & \gc{1.1T} & \gc{218M} & \gc{243\scriptsize{ms}}\\
        \gc{MViTv2-L} & \gc{Sup, \scriptsize{21K}} & \gc{53.6} & \gc{47.5} & \gc{{1.3T}} & \gc{{239M}} & \gc{447\scriptsize{ms}}\\
        MViTv2-L & Sup & 53.2 & 47.1 & \ul{1.3T} & \ul{239M} & 447\scriptsize{ms}\\
        ViTDet-L  & MAE & \textbf{55.6} & \textbf{49.2} & 1.9T & 331M & \ul{396\scriptsize{ms}}\\
        \baseline{\shortname-L} &{MAE} & \ul{55.0} & \ul{48.6} & \textbf{{1.2T}} & \textbf{{236M}}  & \textbf{{340\scriptsize{ms}}}\\   
    \end{tabular}
}
\caption{\textbf{COCO object detection and segmentation} using Mask-RCNN. All methods are following the training recipe from~\citet{vitdet} and pretrained on ImageNet-1K by default. Methods using ImageNet-21K pretraining are \gc{de-emphasized}. Test time is measured on a single V100 GPU with full precision.
}
\label{tab:coco}
\vspace{-10pt}
\end{table}
%##################################################################################################

\section{Conclusion}
In this work, we create a simple hierarchical vision transformer by taking an existing one and removing all its bells-and-whistles while supplying the model with spatial bias through MAE pretraining. The resulting architecture, \shortname{}, is more effective than current work on image recognition tasks and surpasses the state-of-the-art on video tasks. We hope \shortname{} can allow future work to do more, faster.

% \clearpage
\appendix

%%%%%%%%%%%%%%%%%%%%%%%%%%%%%%%%%%%%%%%%%%%%%%%%%%%%%%%%%%%%%%%%%%%%%%%%%%%%%%%
%%%%%%%%%%%%%%%%%%%%%%%%%%%%%%%%%%%%%%%%%%%%%%%%%%%%%%%%%%%%%%%%%%%%%%%%%%%%%%%
% APPENDIX
%%%%%%%%%%%%%%%%%%%%%%%%%%%%%%%%%%%%%%%%%%%%%%%%%%%%%%%%%%%%%%%%%%%%%%%%%%%%%%%
%%%%%%%%%%%%%%%%%%%%%%%%%%%%%%%%%%%%%%%%%%%%%%%%%%%%%%%%%%%%%%%%%%%%%%%%%%%%%%%

\section{Implementation Details}
A mask unit for video corresponds to a block of $2~\text{frames}\times32~\text{px}\times32~\text{px}$ (as opposed to images which use $1\times32\times32$). Following \citet{mae-st}, each token in \shortname{}~on video corresponds to 2 frames of the input. Since the mask units also span 2 frames, the window sizes for Mask Unit Attention do not change for video (i.e., $1\times8\times8$ tokens in the first stage, $1\times4\times4$ tokens in the second stage)---meaning we use exactly the same implementation for images and video (just the mask unit size is changed). We use learned spatial (separable spatio-temporal) position embeddings for images (video). These are all the differences between \shortname{}~for images and for video. The rest of the encoder is completely agnostic to spatio-temporal structure.

As in \citet{maskfeat}, we remove $Q$-pooling before the last stage for MAE pretraining only. This is done so that MAE settings from prior work using ViT also work for \shortname{}~with minimal modifications. This introduces little extra computation as stage 4 is small. If desired, by \textit{design}, pretraining with \shortname{} can also work without removal of query pooling during pretraining, since a mask unit of $1\times8\times8$ tokens would correspond to $1$ distinct token in the last stage.

\subsection{Video Experiments}

\paragraph{Kinetics-400, -600, -700.} Our settings mainly follow \citet{mae-st}. We report the pretraining and finetuning settings for our main results on the Kinetics-400 \cite{k400}, -600 \cite{k600} and -700 \cite{k700} human action datasets in Tab.~\ref{tab:kineticsSettings}. Epochs are always reported as effective epochs \cite{mae-st}, i.e. accounting for repeated sampling. We use $16\times4$ sampling as in \citet{mae-st}.

%##################################################################################################
\begin{table}[t]
\centering
%#################################################
%#################################################
\subfloat[{Pretraining}]{
\centering
\begin{minipage}{1.0\linewidth}{\begin{center}
\tablestyle{3pt}{1.00}
\begin{tabular}{y{115}|x{115}}
config  & value \\
\shline
optimizer & AdamW~{\tiny\cite{adamw}} \\
optimizer momentum & {$\beta_1, \beta_2{=}0.9, 0.95$} \\ %\cite{Chen2020c} \\
weight decay & {0.05} \\
learning rate & 8e-4 ({\footnotesize B, B+, L}); - / 8e-4 / 3.2e-4 (H) \\
learning rate sch. & cosine decay {\tiny\cite{sgdr}}\\
warmup epochs {\tiny \cite{Goyal2017}} & {120} \\
epochs& 800 / 1600 / 3200 \\
repeated sampling~{\tiny \cite{Hoffer2020}} & {4} \\ %~\cite{Hoffer2020}
augmentation & {hflip, crop $[0.5, 1]$} \\
batch size & 512 \\
%gradient clipping & {0.02} \\
num. decoder blocks & 8 \\
num. decoder heads & 8 \\
mask ratio & 0.9\\
drop path~{\tiny\cite{droppath}} &  0.1 (B); 0.2 (B+, L); - / 0.3 / 0.4 (H)\\
\end{tabular}
\end{center}}\end{minipage}
}
\vspace{-5pt}
%#################################################
%#################################################
\subfloat[{Finetuning}]{
\centering
\begin{minipage}{1.0\linewidth}{\begin{center}
\tablestyle{3pt}{1.00}
\begin{tabular}{y{95}|x{130}}
config& value \\
\shline
optimizer & AdamW \\
optimizer momentum & {$\beta_1, \beta_2{=}0.9, 0.999$} \\
weight decay & 0.05 \\
learning rate & 8e-4 (B, B+, L), 4e-4 (H) \\
learning rate schedule & cosine decay \\
warmup epochs & 10 \\
epochs & 150 (B, B+), 100 (L, H) \\
repeated sampling & 2 \\
augmentation & RandAug (7, 0.5)~{\tiny\cite{randaug}} \\
batch size &  256  \\
gradient clipping & 5.0 \\
mixup~{\tiny\cite{mixup}} & 0.8 \\
cutmix~{\tiny\cite{cutmix}} & 1.0 \\
label smoothing~{\tiny\cite{inception}} & 0.1 \\
drop path & 0.2 (B, B+, L), 0.3 (H) \\
dropout~{\tiny\cite{dropout}} & 0.3 (B, B+), 0.5 (L, H) \\
layer-wise decay~{\tiny\cite{electra}} & - / 0.85 / 0.8 (B, B+); 0.925 / 0.9 / 0.875 (L, H)   \\
\end{tabular}
\end{center}}\end{minipage}
}
\vspace{-5pt}
\caption{{\bf Settings for Kinetics-400, -600, -700.} Notation: setting corresponding to 800 / 1600 / 3200 epochs of pretraining.}\label{tab:kineticsSettings}
\end{table}
%##################################################################################################

\paragraph{Something-Something-v2 (SSv2).} We evaluate \shortname{}-L on the SSv2 dataset~\cite{goyal2017something}. SSv2 is a dataset focusing on human-object interaction classification. We pretrain \shortname{}-L on either Kinetics 400 or SSv2 and finetune on SSv2. We report the top-1 classification accuracy in Tab.~\ref{tab:sota:ssv2}.  We provide further details about the pretraining and finetuning settings on SSv2 in Tab.~\ref{tab:ssv2settings}.

%##################################################################################################
\begin{table}[t]
\centering
%#################################################
%#################################################
\subfloat[{Pretraining}]{
\centering
\begin{minipage}{1.0\linewidth}{\begin{center}
\tablestyle{3pt}{1.00}
\begin{tabular}{y{80}|x{65}}
config  & value \\
\shline
optimizer & AdamW \\%{AdamW~\cite{Loshchilov2019}} \\
optimizer momentum & {$\beta_1, \beta_2{=}0.9, 0.95$} \\ %\cite{Chen2020c} \\
weight decay & {0.05} \\
learning rate & 8e-4 \\
learning rate schedule & cosine decay\\%~\cite{Loshchilov2016} \\
warmup epochs & {30} \\ %~\cite{Goyal2017}
epochs& {1600} \\
%repeated sampling & {4} \\ %~\cite{Hoffer2020}
augmentation & {crop $[0.5, 1]$} \\
batch size & 1024 \\
gradient clipping & {0.02} \\
num. decoder blocks & 8 \\
num. decoder heads & 8 \\
mask ratio & 0.9 \\
drop path & 0.2 \\
\end{tabular}
\end{center}}\end{minipage}
}
\vspace{-5pt}
%#################################################
%#################################################
\subfloat[{Finetuning}]{
\centering
\begin{minipage}{1.0\linewidth}{\begin{center}
\tablestyle{3pt}{1.00}
%\scriptsize
\begin{tabular}{y{80}|x{65}}
config& values \\
\shline
optimizer & {SGD} \\
weight decay & {1e-4} \\
learning rate & 0.16 / 0.08 \\
learning rate schedule & cosine decay\\ %~\cite{Loshchilov2016}} \\
warmup epochs & {3} \\ %~\cite{Goyal2017}
epochs & {40} \\
augmentation & RandAug (7, 0.5) \\ %~\cite{Cubuk2020}
batch size & 256 / 128  \\
mixup & 0.8 / - \\ %~\cite{Zhang2018a}
cutmix & 1.0 / - \\ %~\cite{Yun2019}
label smoothing & 0.1 / - \\ %~\cite{inception}
drop path & 0.1 / 0.2 \\ %~\cite{Huang2016}
dropout & 0.5 \\ %~\cite{Srivastava2014}
layer-wise decay & 0.875 \\ %~\cite{Clark2020}
\end{tabular}
\end{center}}\end{minipage}
}
\vspace{-5pt}
\caption{{\bf Settings for SSv2.} Notation: setting corresponding to \shortname{}-L / L$_{32}$.}\label{tab:ssv2settings}
\end{table}
%##################################################################################################

\paragraph{AVA v2.2.} We perform transferring experiments on AVA v2.2~\cite{gu2018ava} for human action localization in video. We adopt a detection framework following~\cite{slowfast} for human action localization. We extract ROI features from the feature map of the last layer in \shortname{} and pool the ROI features via spatial max-pooling. We then use a linear classifier trained with cross entropy loss to predict the action class. We use the center crop for \shortname{} in the evaluation and report the mAP in Tab.~\ref{tab:sota:ava}. We use Kinetics pretrained and finetuned \shortname{} in the experiments. We provide details about the finetuning setting on AVA v2.2 in Tab.~\ref{tab:avasettings}.

\begin{table}[h!]\centering
%\vspace{.5em}
%\subfloat[{\shortname{} fine-tuning on AVA}]{
\tablestyle{3pt}{1.00}
%\scriptsize
\begin{tabular}{y{80}|x{80}}
config& values \\
\shline
optimizer & {SGD} \\
weight decay & {1e-8} \\
learning rate & 3.6\\
learning rate schedule & cosine decay\\ %~\cite{Loshchilov2016}} \\
warmup epochs & {5} \\ %~\cite{Goyal2017}
epochs & {30} \\
batch size & 128  \\
drop path & 0.4 \\ %~\cite{Huang2016}
dropout & {0.5} \\ %~\cite{Srivastava2014}
layer-wise decay & 0.875 \\ %~\cite{Clark2020}
\end{tabular}
%}
\caption{{\bf Settings for AVA.} \shortname{-L} and \shortname{-H} finetuning settings on AVA.}\label{tab:avasettings}
\vspace{-10pt}
\end{table}

\subsection{Image Experiments}
\paragraph{ImageNet-1K.} Our settings mainly follow \citet{mae}. We report the pretraining and finetuning settings for our main results in Tab.~\ref{tab:in1kSettings}.

%##################################################################################################
\begin{table}[t]
% \vspace{-10pt}
\centering
%#################################################
%#################################################
\subfloat[{Pretraining}]{
\centering
\begin{minipage}{1.0\linewidth}{\begin{center}
\tablestyle{3pt}{1.00}
\begin{tabular}{y{90}|x{115}}
config  & value \\
\shline
optimizer & AdamW \\
optimizer momentum & {$\beta_1, \beta_2{=}0.9, 0.95$} \\ %\cite{Chen2020c} \\
weight decay & {0.05} \\
learning rate & 8e-4 \\
learning rate sch. & cosine decay \\
warmup epochs & {40} \\
epochs& 400 / 1600 \\
augmentation & {hflip, crop $[0.2, 1]$} \\
batch size & 4096 \\
num. decoder blocks & 8 \\
num. decoder heads & 16 \\
mask ratio & 0.6\\
drop path &  0.0 (T, S); 0.2 (B, B+, L); 0.3 (H)\\
\end{tabular}
\end{center}}\end{minipage}
}
\vspace{-5pt}
%#################################################
%#################################################
\subfloat[{Finetuning}]{
\centering
\begin{minipage}{1.0\linewidth}{\begin{center}
\tablestyle{3pt}{1.00}
\begin{tabular}{y{90}|x{115}}
config & value \\
\shline
optimizer & AdamW \\
optimizer momentum & $\beta_1, \beta_2{=}0.9, 0.999$ \\
weight decay & 0.05 \\
learning rate & 2e-3 (T, S, B); 1e-3 (B+, L, H) \\
learning rate schedule & cosine decay \\
warmup epochs & 5 \\
epochs & 300 (T); 200 (S); 100 (B, B+); 50 (L, H) \\
augmentation & RandAug (9, 0.5)\\
batch size &  1024  \\
mixup & 0.8 \\
cutmix & 1.0 \\
label smoothing & 0.1 \\
drop path & 0.1 (T, S, B, B+); 0.2 / 0.1 (L); 0.3 (H) \\
layer-wise decay & 0.65 (T, S); 0.7 (B, B+); 0.9 / 0.85 (L); 0.85 (H) \\
\end{tabular}
\end{center}}\end{minipage}
}
\vspace{-5pt}
\caption{{\bf Settings for ImageNet-1K.} Notation: setting corresponding to 400 / 1600 epochs of pretraining.}\label{tab:in1kSettings}
\end{table}
%##################################################################################################

\paragraph{Transfer learning on iNaturalists and Places.} We conduct transfer learning experiments on classification datasets including iNaturalist2017, iNaturalist2018, iNaturalist2019~\cite{van2018inaturalist} and Places365~\cite{NIPS2014_3fe94a00}. Following~\cite{mae}, we adjust learning rate, training epochs on each dataset. We search the layer-wise decay among 0.875, 0.9 and 0.925, drop path rate between 0.1 to 0.5, and the dropout rate among 0.1, 0.2 and 0.3. For \shortname{}-H$_{448}$, we set the learning rate decay of the positional embedding to 0.5 instead of following the layer-wise decay rule.

\paragraph{COCO.} We use the Mask R-CNN~\cite{maskrcnn} framework in Detectron2~\cite{Wu2019} for object detection and instance segmentation experiments on the COCO dataset. Similar to ViTDet~\cite{vitdet}, we use 2 hidden convolution layers for the RPN and 4 hidden convolution layers for the RoI heads for \shortname{} and all comparison detection methods. These layers are followed by LayerNorm layers.
For the training recipe, we follow ViTDet to use input size as 1024$\times$1024 with large-scale jittering (LSJ)~\cite{Ghiasi2021}. We don't use the layer-wise decay during training. Additional hyperparamters can be found in Tab.~\ref{tab:cocosetting}.

\begin{table}[h!]\centering
\tablestyle{3pt}{1.00}
\begin{tabular}{y{100}|x{100}}
config& values \\
\shline
optimizer & {AdamW} \\
optimizer momentum & {$\beta_1, \beta_2{=}0.9, 0.999$} \\
weight decay & {0.1} \\
learning rate & 3.5e-5 (B, B+), 3e-5 (L) \\
learning rate schedule & step-wise decay\\ 
epochs & {100} \\
augmentation & LSJ [0.1, 2.0] \\
batch size & 64  \\
drop path & 0.2 (B, B+), 0.4 (L) \\ 
\end{tabular}
%}

\caption{{\bf Settings for COCO.} \shortname{} finetuning settings on COCO.}\label{tab:cocosetting}
\end{table}

\subsection{Speed Benchmarking}

 We use an NVIDIA A100 40GB GPU, PyTorch v1.12.1 and CUDA 11.4 to benchmark speed for all baselines and our approach, unless otherwise mentioned. Note that we did \textit{not} use Flash Attention \cite{flashattn} or any other attention speed-up mechanism in this paper, though they can be used to further increase speed. For each of the methods, we measure purely the model inference throughput. We compute the throughput with various batch sizes, and report the throughput with the optimal batch size. We use half precision (fp16) to run speed benchmarking unless otherwise specified. We set the input resolution to 224$\times$224$\times$3 for image benchmarking, and 224$\times$224$\times$3 with 16 frames as a clip for video benchmarking. To measure the training time, we measure the speed of a forward-backward pass on a single gpu and extrapolate the total training time according to the size of the dataset and the number of training epochs, ignoring dataloading and communication overheads when training with multiple GPUs.

We report the image benchmarking results on NVIDIA A100 with fp16 and compared with in ViT~\cite{vit}, ConvNextV2~\cite{woo2023convnextv2} and MCMAE~\cite{gao2022mcmae} in Tab.~\ref{tab:speed_a100_fp16}. We provide video benchmarking results in Tab.~\ref{tab:video_speed_a100_fp16}.

\begin{table}[t!]
    \centering
    \tablestyle{4pt}{1.1}
        \begin{tabular}{lcccc}
      size &  ViT & MCMAE  & ConvNextV2 & \shortname \\
      \shline
      T & - & - & 1381 & \bf 2758 \\
      S & - & - & - & \bf 2211 \\
      B & 1448 & 1069 & 646 & \bf 1556 \\
      B+& -    & 936  & -   & \bf 1247 \\
      L & 514  & 381  & 414 & \bf 531  \\
      H & 205  & 194  & 202 & \bf 274  \\
        \end{tabular}
    %}
    \caption{{{\bf Image speed benchmarking} on A100 fp16 (im/s)}. }
    \label{tab:speed_a100_fp16}
\end{table}

\begin{table}[t!]
    \centering
    \tablestyle{15pt}{1.1}
        \begin{tabular}{lcc}
      size &  ViT & \shortname \\
      \shline
      B  & 47.1 & \bf 133.6 \\
      B+ & -    & \bf 84.1  \\
      L  & 17.8 & \bf 40.8  \\
      H  & 11.3 & \bf 20.9  \\
        \end{tabular}
    %}
    \caption{{{\bf Video speed benchmarking} on A100 fp16 (clip/s)}. }
    \label{tab:video_speed_a100_fp16}
\end{table}

\section{From Scratch Supervised Training}
In the main paper, we show that we can replace the spatial biases offered by specialized modules in a hierarchical vision transformer with a strong pretext task like MAE \cite{mae}, thereby \textit{teaching} these spatial biases instead. This renders these bells-and-whistles unnecessary, and we remove them to construct an extremely fast and accurate vision transformer: \shortname{}.

\begin{figure}[t!]
    \centering
    \includegraphics[width=0.99\linewidth]{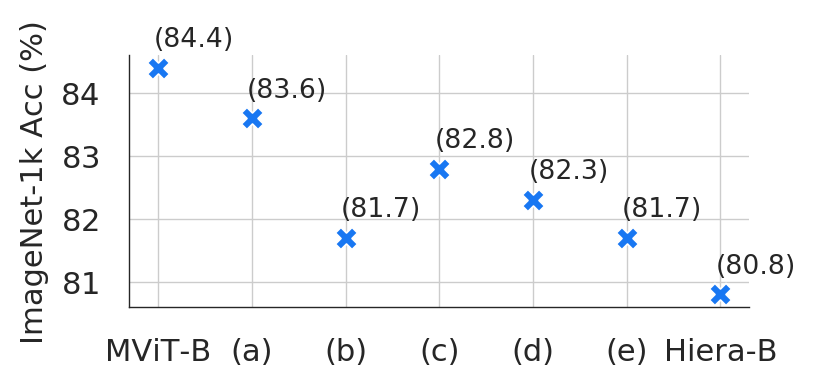}
    \caption{{\bf Training on classification \textit{from scratch}.} Here we repeat the experiment in Tab.~\ref{tab:sparsifying_mvit} but without MAE pretraining, using MViTv2's supervised recipe instead. As expected, the bells-and-whistles that \shortname{}~removes are actually \textit{necessary} when training from scratch---hence their introduction in prior work in the first place. \shortname{}~\textit{learns} spatial biases instead.}
    \label{fig:from_scratch_supervised}
\end{figure}

However, we \textit{do not} claim that these modules are unnecessary \textit{in general}. In fact, here we intend to show the opposite: the reason these spatial biases were necessary in the first place is because they are required when training a vision transformer from scratch with classification. In Fig.~\ref{fig:from_scratch_supervised}, we show this by repeating the ablations in Tab.~\ref{tab:sparsifying_mvit} on ImageNet-1K starting from an MViT-B model and ending at \shortname{-B}, but this time training on classification \textit{from scratch}.

As expected, we see the \textit{opposite} trend as we did when training with a strong pretext task: the bells-and-whistles \textit{are} necessary when training in a classical supervised setting. This reiterates the fact that, by training with MAE, we are \textit{replacing} the need to explicitly build spatial biases into the network's architecture itself.

Note that this also has ramifications for downstream tasks: while prior specialized Vision Transformers like MViT or Swin act like convnets (e.g., you can just use a normal Mask R-CNN \cite{maskrcnn} head for detection), \shortname{}~\textit{acts like a ViT}. Thus, we recommend using transformer-based solutions for downstream tasks such as ViTDet \cite{vitdet} for detection instead of Mask R-CNN.

\clearpage
\bibliography{main}
\bibliographystyle{icml2023}

\end{document}